\documentclass{article}

\usepackage{arxiv}
\usepackage[utf8]{inputenc}
\usepackage{amsmath}
\usepackage[toc,page]{appendix}
\usepackage{fnpct}
\usepackage{float}
\usepackage{dsfont}
\RequirePackage{graphicx}
\RequirePackage{color}
\RequirePackage{verbatim}
\RequirePackage{enumitem}
\usepackage{tabularx}
\usepackage{subcaption}
\usepackage{amsmath}
\usepackage{amstext,amssymb,amsfonts,dsfont}
\usepackage{hyperref}
\usepackage{tabularx}
\usepackage{pgf,tikz}
\usepackage{mathrsfs}
\usepackage{algorithm2e}
\usepackage{algorithmic}
\usepackage[backend=bibtex,style=numeric,natbib=true, maxcitenames=2, maxbibnames=99, maxalphanames=4, backref=true,]{biblatex}
\title{Diverse Counterfactual Explanations for Anomaly Detection in Time Series}

\date{}

\author{
 D\'{e}borah Sulem \\ %
 Department of Statistics\\
 University of Oxford\\
 \texttt{deborah.sulem@stats.ox.ac.uk} \\
  \And
Michele Donini \\
Amazon Web Services \\
 \texttt{donini@amazon.com} \\
 \And
Muhammad Bilal Zafar \\
Amazon Web Services \\
 \texttt{zafamuh@amazon.com} \\
 \And
Fran\c cois-Xavier Aubet \\
Amazon Web Services \\
 \texttt{aubetf@amazon.com} \\
 \And
Jan Gasthaus \\
 Amazon Web Services \\
 \texttt{gasthaus@amazon.com} \\
 \And
Tim Januschowski \\
Zalando Research \\
\texttt{tim.januschowski@zalando.de} \\
 \And
Sanjiv Das \\
 Amazon Web Services  \\ Santa Clara University \\
 \texttt{sanjivda@amazon.com} \\
 \And
Krishnaram Kenthapadi \\
Fiddler AI \\
krishnaram@fiddler.ai \\
\And
C\'{e}dric Archambeau \\
Amazon Web Services \\
 \texttt{cedrica@amazon.com} \\
}

\begin{document}

\maketitle

\begin{abstract}

Data-driven methods that detect anomalies in times series data are ubiquitous in practice, but they are in general unable to provide helpful explanations for the predictions they make. %
In this work we propose a model-agnostic algorithm that generates \emph{counterfactual ensemble explanations} for time series anomaly detection models. Our method %
generates a set of diverse counterfactual examples, i.e, multiple perturbed versions of the original time series that are not considered anomalous by the detection model. Since the magnitude of the perturbations is limited, these counterfactuals represent an ensemble of inputs similar to the original time series that the model would deem normal.
Our algorithm is applicable to any differentiable anomaly detection model.
We investigate the value of our method on univariate and multivariate real-world datasets and two deep-learning-based anomaly detection models, under several explainability criteria previously proposed in other data domains such as \emph{Validity, Plausibility, Closeness} and \emph{Diversity}. We show that our algorithm can produce ensembles of counterfactual examples that satisfy these criteria and thanks to a novel type of visualization, can convey a richer interpretation of a model's internal mechanism than existing %
methods.
Moreover, we design a sparse variant of our method %
to improve the interpretability of counterfactual explanations for high-dimensional %
time series anomalies. In this setting, our explanation is localized on only a few dimensions and
can therefore be communicated more efficiently to the model's user.

\end{abstract}


\section{Introduction}

Anomaly detection in time series is a common data analysis task that %
can be defined as identifying outliers, i.e., observations that do not belong to a reference distribution and call for further investigation.
For instance, anomaly detection %
is leveraged to localize a defect in %
computing systems, disclose a fraud in financial transactions, or diagnose a disease from health records \cite{blazquezgarcia2020review}. %
However, taking appropriate actions in the presence of anomalies typically requires an understanding of why the model marked the data as anomalous.
Therefore, providing explanations for models that detect anomalies can guide %
the decision process of their users and provide a justification for a chosen action. In spite of its %
practical relevance, explaining anomalies detected by ML models is still an understudied problem, all the more in the challenging setting of multivariate time series data.

More precisely, %
a model that detects anomalies classifies each timestamp of a time series as  anomalous or not. Common examples of anomalies are spike outliers, change-points or drifts \cite{blazquezgarcia2020review}. However, in practice %
anomalies cannot be easily categorised and require \textit{ad hoc} analysis. %
Moreover, several  state-of-the-art models %
involve complex deep learning (DL) %
classifiers, such as LSTMs \cite{Malhotra2015LongST}, RNNs \cite{audibert2020} or TCNs \cite{bai2018empirical, carmona2021neural}, %
whose internal mechanisms to detect anomalies are %
opaque. This lack of transparency can prevent these models from being deployed in consequential contexts \cite{brown2018recurrent,bhatt2020explainable}. To improve their comprehensibility, prior work has proposed to include interpretable blocks in machine learning models (e.g., attention mechanism in RNNs \cite{brown2018recurrent}) or design model-specific explainability methods 
(e.g., feature-importance scores for Isolation Forests \cite{carletti2021interpretable}).

In the larger spectrum of methods for explaining time series models, most existing ones consist of feature-saliency estimation \cite{crabbe2021explaining, Pan2020SeriesST}, which attributes scores to features in terms of their relative contribution to the model's prediction. Although these techniques have provided valuable insight in image classification tasks \cite{fong2019understanding}, it is often a weak form of explanation for anomalies in time series since it essentially indicates that the features of the anomalous subsequence are salient.
In Figure \ref{fig:DM}, we show an example of applying the Dynamic Masks method \cite{crabbe2021explaining} to explain an anomaly prediction score. The saliency scores output by this method are high on the timestamps containing the anomalous observations (highlighted in red in Figure \ref{fig:OTS}) and on few timestamps preceding the anomaly. In particular, they do not explain \emph{why} the model classified the input time series as anomalous (or not) and \emph{what} it has learnt as the normal data distribution.
In practice, a user of an anomaly detection model is interested in (a) knowing what can be changed in the input data to avoid encountering the anomaly again in the future (preferentially with minimal cost), and (b) understand the model's sensitivity to a particular %
anomaly.

With these properties in mind, we aim to generate counterfactual samples for explaining anomalies. 
Counterfactual explanations have been previously proposed for time series classifiers \cite{Ates_2021, delaney2021instancebased, Karlsson1379883} and can achieve criterion (a). A counterfactual example (or for short, a counterfactual) is an instance-based explanation in the form of a perturbed input on which the model's prediction value is different from the original model output. It thus indicates what modifications of the input must be made to obtain a dissimilar prediction.   
It is often defined as an instance $X’$ minimizing a cost function such as \cite{wachter2018counterfactual}:
\begin{align*}
    L(X,X',y',\lambda) = \lambda (f(X') - y')^2 + d(X,X'),
\end{align*}
where $f$ is the prediction model, $X$ is the original input, $y'$ is a desired output value  (e.g. a different predicted label in classification contexts), 
$d(.,.)$ is an appropriate distance on the input space and $\lambda$ is a trade-off parameter. Counterfactual explanations have notably been utilized for models that accept or reject loan applications as they provide actionable feedback on personal records \cite{verma2020counterfactual}. %
In their basic definition, they are closely related to adversarial examples \cite{verma2020counterfactual}, however, their properties and their utility are distinct. Adversarial examples are often weakly constrained and used as hard instances to train more robust models, whereas counterfactuals are post-hoc explanations that need to be plausible examples to be interpretable. 

Counterfactual explanations are particularly useful for explaining anomalies in time series detected by a given model, in which case
a counterfactual is another time series (or subsequence) that does not contain observations detected as anomalous and is similar to the anomalous subsequence.
It thus corresponds to the closest normal or expected behaviour according to the model. For example, if the anomalous time series is a temporal record of the blood glucose level of a patient at risk, a counterfactual could be an alternative series of values contained in a non-critical interval. Hence, counterfactual explanations can reveal the boundaries of the normal time series distribution according to the model and, consequently, its sensitivity to possible issues. Unfortunately, as often observed in distinct contexts \cite{russell2019efficient}, a single counterfactual is only a partial explanation, satisfying a particular trade-off between predefined criteria. One general solution is therefore to provide an \emph{ensemble} of diverse counterfactuals \cite{russell2019efficient, Mothilal_2020, Dandl_2020}. However, for time series anomaly detection, there is no existing approach for generating these ensembles, nor a strategy to effectively communicate these more complex explanations to the model's user for general data types.   

In this work, we make the following contributions in this under-explored domain: %
\begin{itemize}
    \item We introduce a model-agnostic method that generates \emph{counterfactual ensemble explanations} for anomalies detected by an algorithm in a univariate or multivariate time series. Our ensemble explanation is a set of counterfactual examples that can be used individually, or analysed together to investigate the model's sensitivity and the range of perturbation that can be applied to the original input. 
    
    \item We propose an interpretable visualization of the counterfactual ensemble
    explanation, in particular by associating the counterfactual examples and their prediction scores under the model.
\end{itemize}

Figure \ref{fig:method} summarizes our propositions, in contrast to existing explainability methods for time series models. The advantages of our explanation is to be (a) \emph{sparse}, in the sense that it is localized on a few time series features (b) \emph{optimal} in that it minimally modifies these features and (c) \emph{rich} by diversifying the possible perturbations. Our method is applicable to any differentiable anomaly detection model %
and can be delineated into two variants, whose respective uses depend on prior knowledge of the data distribution. %
\emph{Dynamically Perturbed Ensembles (DPEs)} leverage dynamic perturbation operators \cite{crabbe2021explaining}, which induce a modification of a time series according to a pre-defined mechanism. %
\emph{Interpretable Counterfactual Ensembles (ICEs)} do not apply such a perturbation mechanism and can be applied in absence of domain expertise.
Additionally, we design for high dimensional time series a sparse version of our method that %
provides a more parsimonious and readable explanation.

After succinctly reviewing existing work in the field of model explainability for time series and counterfactual explanations in Section~\ref{sec:related_work}, we describe the general set-up %
in Section \ref{sec:setup}. In Section \ref{sec:methodology}, we present the technical details of our method. Then in Section \ref{sec:experiments}, we demonstrate the effectiveness of our method on DL-based detection models and benchmark datasets. 
We notably adapt metrics that have been previously used for counterfactual explanations in distinct data domains, such as \emph{Validity, Plausibility, Closeness} and \emph{Diversity} \cite{Mothilal_2020, rodriguez2021trivial, karimi20a}. Moreover, we qualitatively show on several case studies that our novel visualization can provide a comprehensible insight into the model's local decision boundary and sensitivity. Finally, we conclude in Section \ref{sec:discussion} with a summary of our results and possible future developments.

\section{Related work}\label{sec:related_work}

Explainability methods %
for users of machine learning models have developed along two paradigms: building models with interpretable blocks or designing model-agnostic methods that can be applied to any model already deployed. For time series data, RETAIN \cite{choi2017retain} incorporates an attention-mechanism in an RNN-based model
while Dynamic Masks \cite{crabbe2021explaining} is a model-agnostic algorithm that produces sparse feature-importance masks on time series using dynamic perturbation operators. In fact, many methods for time series adapt algorithms designed for tabular or image data: %
for instance, TimeSHAP \cite{Bento_2021} extends SHAP, a feature-attribution method that approximates the local behaviour of a model with a linear model using a subset of features. 
Another interesting line of work %
interprets CNNs for time series models using Shapelet Learning \cite{MaZLHC20}. Shapelets are subsequences that are learnt from a dataset to build interpretable time series decompositions.

Nonetheless, previously cited work for time series are feature-saliency estimation methods. Although they are notably helpful to localize the important parts of time series (in terms of their contribution to the model's prediction), they can only weakly explain anomaly detection models.
Moreover, \emph{instance- or example-based} explanations can be more easily interpreted by a non-expert person \cite{wachter2018counterfactual}.  %
These methods explain a prediction on a single instance by comparing it to another real or generated example, e.g., the most typical examplar of the observed phenomenon (a \emph{prototype} \cite{hautamaki}) or a contrastive examplar related to a distinct behaviour  (a \emph{counterfactual} \cite{Ates_2021, delaney2021instancebased, Karlsson1379883}).
For time series classifiers,  counterfactuals can be generated by swapping the values of the most discriminative dimensions %
with those from another training instance~\cite{Ates_2021}. %
Unfortunately, this approach can yield implausible subsequences, that do not belong to the data manifold \cite{carletti2021interpretable}, e.g., by breaking correlations between the dimensions of multivariate time series. The Native Guide algorithm \cite{delaney2021instancebased} does not suffer from the previous issue but uses a perturbation mechanism on the Nearest Unlike Neighbor in the training set using the model's internal feature vector. Lastly, for a k-NN and a Random Shapelet Forest classifiers, \cite{Karlsson1379883} design a tweaking mechanism to produce counterfactual time series. %

However, these methods %
necessitate knowledge of the model's internal mechanism and/or access to its training dataset, which can be expensive. Additionally, these counterfactual explanations suffer from the so-called \emph{Rashomon effect} \cite{molnar2019}, i.e., the fact that several equally-good perturbed examples might exist and be informative for the model's user. In this case, one might benefit from knowing multiple ones, before choosing the most helpful example in a specific context \cite{Mothilal_2020}. 
For linear classifiers of tabular data, a set of diverse counterfactuals can be obtained by sequentially adding constraints along the optimization iterations of the perturbation algorithm~\cite{russell2019efficient}, whereas  %
the Multi-Objective Counterfactuals algorithm \cite{Dandl_2020} records multiple perturbed examples generated along the iterations of a genetic algorithm. These counterfactual sets therefore contain different trade-offs between conflicting criteria.
While in the previous methods, diversity is not explicitly enforced, the DiCE algorithm \cite{Mothilal_2020} includes a penalization on counterfactuals' similarity based on Determinantal Point Processes. %
In a similar fashion, for image classifiers, DiVE \cite{rodriguez2021trivial} perturbs the latent features in a Variational Auto Encoder and penalises pairwise similarity between perturbations, while \cite{karimi20a} propose a general framework for generating counterfactual examples with  diversity constraints in heterogeneous data.

Nevertheless, to our knowledge, there is no method to provide diverse counterfactual explanations for time series data, %
 \textit{a fortiori} in the context of anomaly detection. 
 Moreover, previous works that have enriched counterfactual explanations with diverse examples have not discussed the additional challenge of communicating efficiently %
a set of instances rather than a single one. 
Before exposing the technicalities of our method, we describe the general set-up in the next section.

\section{General set-up}\label{sec:setup}

In this work, we assume that anomalies in a time series are unpredictable and out-of-distribution subsequences.
Hence, an anomaly is a significant deviation from a given reference behaviour. In the remainder, we will not make a %
distinction between anomaly, outlier and anomalous/abnormal/atypical observation. %
Not-anomalous data points will be considered as belonging to the data distribution, and denoted as the reference/normal/typical/expected behaviour. We will also refer to the latter as the \emph{context}.  

For the description of the general set-up, we introduce the following notations: for an integer $k \in \mathbb{N}$, $[k]$ denotes the set $\{i; \: 1 \leq i \leq k\}$ and for $x \in \mathbb{R}$, let $x_+ = \max(0,x)$. For a vector $v \in \mathbb{R}^n$, we denote $v_i$ its $i$-th coordinate and for $X \in \mathbb{R}^{m \times n}$ a matrix or multivariate time series, $X_i$ denotes respectively the $i$-th row or the $i$-th observation.

\paragraph{Anomaly detection model}

We assume that we are given an anomaly detection model which we can use to predict -- or rather in this context, detect -- anomalies on a time series of any given length. We consider a general setting where time series are multivariate  and the model processes all dimensions (or \emph{channels}) jointly.
More precisely, we denote $X \in \mathbb{R}^{T \times D}$  %
a time series with $T$ timestamps and $D$ dimensions. %
The prediction function of the model, denoted by $f$, is used to classify each timestamp $t \in [1,T]$ of $X$ as "anomalous" (i.e., label 1) or "not-anomalous" (i.e., label 0).  In fact, the prediction $f(X) \in \mathbb{R}^{T}$ is a vector of anomaly scores for each timestamp (e.g., probability scores of being anomalous)  which transforms into a vector of 0-1 labels using the model's classification rule (e.g. a threshold on these scores). Note that the dimension of the vector $f(X)$ might be smaller than $T$ if the model needs a warm-up interval. %

In practice, these models often detect anomalous timestamps by subdividing time series into smaller time windows %
and classifying the latter (therefore each timestamp or a subset of them in these sub-windows). In other works, to output a prediction on a single timestamp, the "receptive field" of a model is generally a fixed-size (typically small) window. Let's denote $W \in \mathbb{R}^{L \times D}$ a window of size $L$ and consider the following general set-up: the window $W = [W_C, W_S]$ is subdivided by the model into two parts, with $W_C  \in \mathbb{R}^{(L-S) \times D}$ a \emph{context} part (that can be empty if the context is implicit once the model is trained) and $W_S \in \mathbb{R}^{S \times D}$ a \emph{suspect} part, for which the model makes a prediction. More precisely, $f(W) \in \mathbb{R}^S$ is the anomaly score of the window $W_S$ and, without loss of generality, we suppose that $f(W) \in [0,1]^S$.
We also denote $\theta \in [0,1]$ the anomaly detection rule, i.e., a label 1 is given to $W_S$ if for some $i \in [S]$, $f(W)_i > \theta$.

Examples of anomaly detection models with the previously described mechanism are NCAD \cite{carmona2021neural}, where the context window has typically thousands of timestamps and the suspect window has 1 to 5 timestamps, and USAD \cite{audibert2020}, where $W = W_S$ and $L = 5$ or $10$. In the latter case, the context is implicit and the whole training set is considered as normal data and thus the context of anomalies detected in a test time series. %

\begin{figure}
\centering
\begin{subfigure}[b]{1.\textwidth}
     \centering
    \includegraphics[width=0.8\textwidth,trim={1cm 0.5cm 1cm 0cm}, clip]{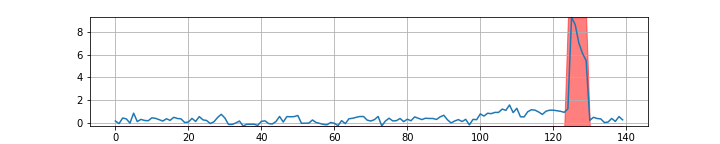}
    \subcaption{Original time series window}
    \label{fig:OTS}
\end{subfigure}
\hfill
\vspace{-0.7cm}
\begin{subfigure}[b]{1.\textwidth}
\centering
\hspace{0.5cm}
    \includegraphics[width=0.7\textwidth,trim={4cm 0.35cm 5.5cm 0cm},clip]{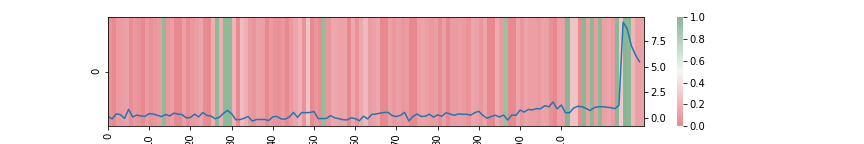}
    \subcaption{Feature-saliency map}
    \label{fig:DM}
\end{subfigure}
\hfill
\vspace{-0.7cm}
\begin{subfigure}[b]{\textwidth}
          \centering
          \hspace{0.1cm}
          \includegraphics[width=0.77\textwidth,trim={1.7cm 0.45cm 1cm 0cm},clip]{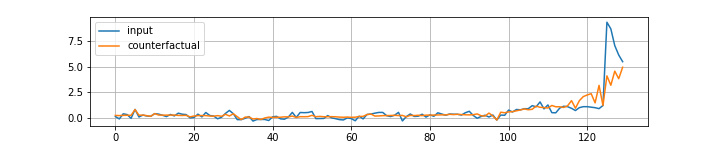}
   \subcaption{Counterfactual example}
   \label{fig:CTS}
\end{subfigure}
\hfill
\vspace{-0.7cm}
\begin{subfigure}[b]{\textwidth}
\hspace{0.0cm}
          \centering
 \includegraphics[width=0.78\textwidth, trim={2cm 0.5cm 5.5cm 0cm},clip]{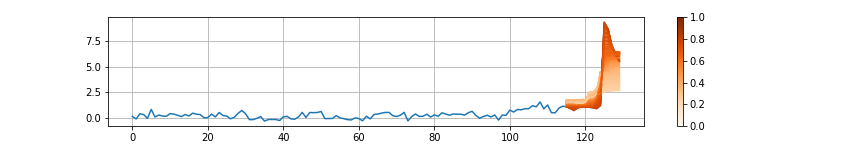}
   \subcaption{Our counterfactual ensemble explanation}
   \label{fig:OE}
\end{subfigure}
\hfill
\vspace{-0.7cm}
\begin{subfigure}[b]{\textwidth}
\hspace{0.0cm}
          \centering
 \includegraphics[width=0.78\textwidth, trim={2cm 0.1cm 5.5cm 0cm},clip]{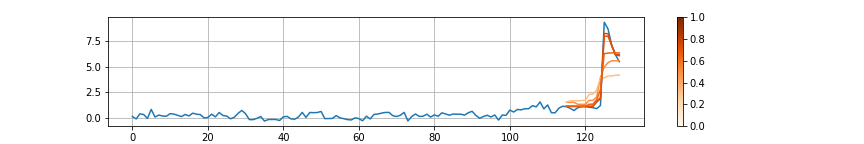}
   \subcaption{Counterfactual examples from our ensemble explanation}
   \label{fig:OEex}
\end{subfigure}
    \caption{Comparison between existing explainability methods for time series and ours, in the context of anomaly detection. The original input (\ref{fig:OTS}) is a univariate time series window containing an anomalous subsequence (highlighted in red). Explanations from a feature-importance method (Dynamic Masks \cite{crabbe2021explaining}) (\ref{fig:DM}), an instance-based method (counterfactual example) (\ref{fig:CTS}) and our method (\ref{fig:OE}) are represented. In \ref{fig:DM}, the salient timestamps have saliency scores closed to one and are highlighted in green. In \ref{fig:OE}, all the examples from our counterfactual time series ensemble are plotted on the anomalous sub-window (\ref{fig:OE}), and the orange color map indicates their anomaly probability scores (between 0 and 1) given by the model. In \ref{fig:OEex}, we also plot five examples from this ensemble.}
    \label{fig:method}
\end{figure}

\paragraph{Counterfactual explanation}
In most cases, a single anomaly is a short subsequence, and can therefore be contained in one or few contiguous subwindows $W_S$. For ease of exposition, we suppose that an anomaly is contained in one suspect window. An example is shown in  Figure \ref{fig:OTS} where a suspect window $W_S$ (highlighted in red) contains an anomaly. A counterfactual example for model $f$ detecting an anomaly in  $W_S$ (i.e., for some $i \in [S]$,  $f(W)_i > \theta $), is an alternative window $\widetilde{W} = [W_C, \widetilde{W}_S]$ such that all predicted labels are 0 (i.e., for any $i \in [S]$, $f(\widetilde{W})_i < \theta $). Since the context of the anomaly 
is also key to its detection by the model,
and if $W$ does not contain a context window $W_C$, we choose to add in the counterfactual example $\widetilde{W}$ a fixed size window $W_C$, that immediately precedes $W$ in the time series. Note that we implicitly suppose that anomalies are not too close to each other so that the additional context window does not contain any anomaly. With a slight abuse of notations, we still denote $\widetilde{W}$ the obtained counterfactual example.

\paragraph{Properties of counterfactual explanations}

There are four largely consensual properties that convey value and utility to counterfactual explanations in the context of model elicitation \cite{verma2020counterfactual}:
\begin{enumerate}
    \item  \emph{Validity} or \emph{Correctness}: achieving a desired model output, e.g. changing the predicted class label in classification; this is the key goal of a contrastive explanation.
    \item \emph{Parsimony} or \emph{Closeness}: minimally and sparsely changing the original input; this is motivated by practical feasibility of the counterfactual if the input features are actionable, and by readibility of the information communicated to the model's user.
     \item \emph{Plausibility}: counterfactual explanations need to contain realistic examples of normal subsequences. %
    \item Being computable within a reasonable amount of time and with acceptable computing resources. %
\end{enumerate}

In the context of an anomaly detected in a time series, property (1) is equivalent to flipping the anomaly detection model's prediction label %
from 1 %
to 0 %
(i.e., achieving a anomaly prediction score below the classifier threshold). Property (2) can be enforced by restricting the perturbation of the input on a small window containing the anomaly (i.e., the suspect window $W_S$) and on few dimensions of the time series (if the anomalous features are only located on some channels). %
Property (3) requires that the counterfactual belongs to the normal data distribution. If the latter is not known or estimated, this criterion can be complicated to evaluate, but some prior knowledge such as the time series' regularity, seasonality, or bounds can be leveraged. Property (4) potentially depends on the specific setting, in particular the cost of using the model's prediction function or its gradient, and the size of the dataset. However, in our context, we assume that accessing the training set of the detection model is particularly expensive, since the latter often decomposes the time series into small windows, leading to a large number of actual training inputs for long time series. %

Unfortunately, those properties are often conflicting (e.g. parsimony and plausibility in the context of a spike outlier), therefore a single counterfactual example can only achieve a particular trade-off between them. In the next paragraph, we motivate the use of counterfactual \emph{ensembles} (or sets) as more comprehensive explanations. %

\paragraph{Diversity as an additional property}

A classical counterfactual explanation is a single counterfactual example that combines properties (1)-(4). However, the best trade-off %
might depend on the particular anomaly or user's range of action. In absence of this prior knowledge, previous work \cite{Mothilal_2020, russell2019efficient} has added the notion of diversity, or range of perturbation, in the list of informative criteria. In particular, it can increase the likelihood of finding a helpful explanation \cite{rodriguez2021trivial}. We also argue that this additional complexity should be adequately communicated to the explanation's recipient, e.g., with a suitable visualization.
An example of our proposed representation is shown in Figure \ref{fig:OE}: the range of time series values and prediction scores spanned by the different counterfactuals in our ensemble explanation effectively informs the user of the possible perturbations and the model's sensitivity.

\section{Methodology}\label{sec:methodology}

In this section, we present our method to generate a counterfactual ensemble explanation for differentiable models, as well as their sparse versions for high-dimensional time series. Additionally, we propose a gradient-free sampling algorithm, called Forecasting Samples (FS), that is applicable to black-box models and relies on a auxiliary probabilistic forecasting model.

\subsection{Gradient-based counterfactual ensemble explanations}\label{sec:gradient_methods}

Most counterfactual algorithms (e.g. Native Guide \cite{delaney2021instancebased}, Growing Spheres \cite{laugel2017inverse}, DiCE \cite{Mothilal_2020}) %
rely on adequately perturbing the input $W$ and optimise the perturbation to enforce some properties of the perturbed example. In our method, using the notations of Section \ref{sec:setup}, we %
first define a penalized objective function over a single counterfactual example $\widetilde{W} = [W_C, \widetilde{W_S}]$,
then use a gradient-descent algorithm to minimize it. The ensemble of examples is built along the optimization path by collecting adequate perturbations. We define two variants of our method: the first one, called \emph{Interpretable Counterfactual Ensemble} (ICE), is an end-to-end method that does not require any domain knowledge input; the second one, \emph{Dynamically Perturbed Ensemble} (DPE), uses an explicit dynamic perturbation mechanism \cite{crabbe2021explaining}. 

\paragraph{Interpretable Counterfactual Ensemble (ICE)}
In this variant, the loss function on a counterfactual example is defined as follows: %
\begin{align}\label{eq:objective_icod}
  \mathcal{L}_{ICE}(\widetilde{W}) =  \mathcal{L}_{pred} (\widetilde{W})  + \mathcal{L}_{c} (\widetilde{W}) + \mathcal{L}_{s} (\widetilde{W}),
\end{align}
where the first term accounts for the Validity property via a hinge loss on the prediction score on $\widetilde{W}$, i.e.,
\begin{align*}
   \mathcal{L}_{pred} (\widetilde{W}) = (f(\widetilde{W})-c)_+,
\end{align*}
with
$c \in [0,1]$ is a margin parameter. %
The second term in \eqref{eq:objective_icod} enforces the Closeness constraint via a penalty similar to the elastic net \cite{zhou05}, here using the Frobenius and the $L_1$ matrix distances: %
\begin{align*}
   \mathcal{L}_{c} (\widetilde{W}) =   \frac{\lambda_1}{S\sqrt{D}} \|\widetilde{W} - W\|_1 + \frac{\lambda_2}{SD} \|\widetilde{W} - W\|_F ,
\end{align*}
where $\lambda_1, \lambda_2 > 0$ are regularization parameters. Finally, the third term of \eqref{eq:objective_icod} enforces Plausibility through temporal smoothness:
\begin{align*}
    \mathcal{L}_{s} (\widetilde{W}) = \frac{\lambda_T}{(S-1)D}   \sum_{i = 1}^{D} \sum_{t=1}^{S-1} |[\widetilde{W}_S]_{(t+1)i} - [\widetilde{W}_S]_{ti}|,
\end{align*}
with $\lambda_T > 0$. The assumption behind this constraint is that normal time series are not too rough and smoother than abnormal windows, therefore realistic perturbations should also be quite smooth.

\paragraph{Dynamically Perturbed Ensemble (DPE)}
Contrary to ICE, this variant uses an explicit perturbation mechanism based on a dynamic perturbation operator \cite{crabbe2021explaining} and a map that spatially and temporally modulates this perturbation. More precisely, a map is a matrix $M \in [0,1]^{S \times D}$ that accounts for the amount of change applied to a timestamp and a dimension in the suspect window $W_S$. A value close to 1 in $M$ indicates a big change %
while a value close to 0 indicates a small change. Our dynamic perturbation operator is a Gaussian blur %
which takes as input a time series window $W$, a timestamp $t \in [L-S,L]$, a dimension $i \in [D]$ and a weight $m \in [0,1]$, and is %
defined as:
\begin{align*}
    \pi_G(W,t, i,m) = \frac{\sum_{t'=1}^L  W_{t'i} \exp(-(t-t')^2/2(\sigma_{max}(1-m))^2)}{\sum_{t'=1}^L \exp(-(t-t')^2/2(\sigma_{max}(1-m))^2)},
\end{align*}
with $\sigma_{max} \geq 0$, a hyperparameter  tuning the blur's temporal bandwidth.  %
We note that the bigger this parameter is, the larger is the smoothing effect of the perturbation. The latter is called \emph{dynamic} in the sense that it modifies a timestamp using its neighbouring times. We also refer to \cite{crabbe2021explaining} for more examples of dynamic perturbation operators.

Finally, for a given map $M$, a perturbed suspect window is given by $[\widetilde{W}_S(M)]_{ti} = \pi(W,L-S+t,i, 1 - M_{ti})$, $t \in [S], i\in [D]$. The loss function is then written in terms of the perturbation map as:
\begin{align}\label{eq:objective_dynamap}
    \mathcal{L}_{DPE}(M) = \mathcal{L}_{pred} (\widetilde{W}(M))  + \frac{\lambda_1}{S \sqrt{D}} \|M\|_1 + \frac{\lambda_2}{S D}  \|W - \widetilde{W}(M)\|_F +  \frac{\lambda_T}{(S - 1)D}  \sum_{i = 1}^{D} \sum_{t = 1}^{S-1} |M_{(t+1)i} - M_{ti}|,
\end{align}
where the first term is the hinge loss, %
and the second and fourth terms account for the sparsity and smoothness constraints, in this case applied on $M$ rather than $\widetilde{W}$ as in \eqref{eq:objective_icod}.

\paragraph{Optimization}
We initialize the counterfactual $\Tilde{W}$ at $W$ and minimize the objective function \eqref{eq:objective_icod} or \eqref{eq:objective_dynamap} using a Stochastic Gradient Descent (SGD) algorithm. Along the iterations of the latter, if we find an example $\widetilde{W}$ such that $\forall i \in [S], f(\widetilde{W})_i < \theta$, %
we add $\widetilde{W}$ to a set $I$. At the end of the iterations, we subsample $N$ counterfactuals from the set $I$ to obtain a diverse counterfactual ensemble. For simplicity, the subsampling scheme is a regular grid over the generation rank of the examples.

\subsection{Sparse counterfactual explanations for high-dimensional time series}

In high-dimensional settings (typically $D > 10$), restricting the perturbations to act on short temporal windows (less than 10 timestamps) is not enough to obtain an interpretable explanation if the counterfactual ensemble explanation spans all dimensions. In fact, the model's user is likely to prefer explanations that are low-dimensional since they are easier to visualize and allow to change a minimal number of system units if the features are actionable. Besides, it is often the case in multi-object systems that an anomaly only affects few dimensions (e.g. a small subsample of monitoring metrics taking abnormal values in a servers network \cite{su2019robust}), hence its explanation should also reflect this low-dimensional property.
For this purpose, we design a sparse version of our gradient-based method that constraints the counterfactual ensemble explanation to be \emph{spatially} sparse (i.e. sparse in the perturbed dimensions, therefore parsimonious in dimensions).

\paragraph{Sparse ICE} In the sparse version of ICE, we restrict the number of perturbed dimensions by introducing a vector $w \in [0,1]^D$ and a matrix $Z \in \mathbb{R}^{S \times D}$, and defining $\widetilde{W}_S(w,Z) =  (w \otimes \mathbf{1}) \odot Z + ((1 - w) \otimes \mathbf{1}) \odot W_S$. The role of $w$ is to select the dimensions in $W_S$ that are perturbed with $Z$. We then consider an objective function in terms of $(Z,w)$:
\begin{align}\label{eq:objective_icod_sparse}
    \mathcal{L}_{ICE, SP}(w, Z) = (f(\widetilde{W}(w,Z))-c)_+ + \frac{\lambda_1}{\sqrt{D}} \|w\|_1 + \frac{\lambda_2}{S D}  \|W - \widetilde{W}(w,Z)\|_F + \frac{\lambda_T}{(S-1)D}   \sum_{i = 1}^{D} \sum_{u = 1}^{S-1} |Z_{(u+1)i} - Z_{ui}| .
\end{align}
Contrary to \eqref{eq:objective_icod}, where the sparsity penalization is applied globally (i.e., both temporally and spatially), the previous objective enforces spatial sparsity through the $L_1$-penalisation on $w$. Another way to see that is to re-interpret objective \eqref{eq:objective_icod} as objective \eqref{eq:objective_icod_sparse} with $w = (1,1, \dots, 1)$, $Z = \widetilde{W}_S$ and replace the $L_1$-penalisation on $w$  by $\frac{\lambda_1}{S\sqrt{D}} \|Z - W_S\|_1$.

\paragraph{Sparse DPE} We apply the same idea to the DPE variant by enforcing the perturbation maps to be spatially sparse. More precisely, we define $M(w,t) = t \otimes w$ with $w \in [0,1]^D$  and $t \in [0,1]^T$ and a loss function in terms of $(w,t)$:
\begin{align}\label{eq:objective_dynamap_sparse}
      \mathcal{L}_{DPE, SP}(w, t) = (f(\widetilde{W}(w,t))-c)_+ + \frac{\lambda_1}{\sqrt{D}}  \|w\|_1 + \frac{\lambda_2}{S D}  \|W - \widetilde{W}(w,t)\|_F + \frac{\lambda_T}{S-1}  \sum_{u=1}^{S-1} |t_{u+1} - t_u| .
\end{align}
Here the smoothness constraint is applied on $t$ to guarantee that $M$ is also smooth in the temporal dimension.

\subsection{Gradient-free approach: Forecasting Set}\label{sec:forecasting_set}

If the anomaly detection model is non-differentiable, we propose an alternative approach that generates a counterfactual ensemble explanation using an appropriate sampling mechanism. %
Machine learning models for time series data sometimes rely on sampling in the context of probabilistic forecasting. Here, we will train an auxiliary probabilistic forecasting method and use it as a generative model of counterfactual subsequences. More precisely, given an input window $W_C \in \mathbb{R}^{L-S \times D}$, our auxiliary model $g$ outputs a distribution over a forecast horizon of $S$ timestamps, $g(W_C)$, from which one can sample forecasting paths. We therefore sample $N$ windows $W_F^{(i)} \sim g(W_C), \: i \in [N]$, then select the ones that are not anomalous according to the anomaly detection model, i.e., our counterfactual ensemble is given by:
$$I_{FS} = \{W_F^{(i)}; \: i \in [N] \: \text{st} \: \forall t \in [S], f([W_C,W_F^{(i)}])_t < \theta\}$$
Intuitively, since the probabilistic forecasting model is trained to learn the data distribution, it generates realistic forecast samples. %
However, the sampling model is oblivious to the original input $W_S$ and therefore the forecasting samples are not restricted to be minimally distant from it.
In Section \ref{sec:experiments}, we will construct and evaluate this approach with a Feed Forward Neural Network (FFNN) for univariate data and a DeepVAR model \cite{Salinas2019HighDimensionalMF} for multivariate data from the GluonTS package~\citep{gluonts} \footnote{https://ts.gluon.ai/index.html}. %

\section{Experiments}\label{sec:experiments}

In this section, %
we test and compare the performances of our method %
on two differentiable models, and the relative advantages of its variants (i.e., ICE, DPE, FS, Sparse ICE, Sparse DPE) in multiple contexts. For this analysis, we have considered two DL anomaly detection models, NCAD \cite{carmona2021neural} and USAD \cite{audibert2020}, %
and four benchmark time series datasets. %
We report in Section \ref{sec:qualitative} a qualitative evaluation of our counterfactual ensemble explanations and their visualization, and in Section \ref{sec:exp_univariate}, a quantitative analysis under the previously defined criteria. Note that this study does not include a comparison to existing baselines, since counterfactual ensemble explanations have not been previously considered for time series data. Although some algorithms such as DiCE \cite{Mothilal_2020} exist in the context of tabular data, we do not deem appropriate to use it in our context since perturbation methods are adapted to each data domain \cite{crabbe2021explaining}. Nonetheless, for the sake of comparison, we also include a naive baseline, which mechanism is described in Section \ref{sec:exp_setup}. Section \ref{sec:exp_metrics} and Section \ref{sec:hp_selection} provide additional details on the explainability metrics and the hyperparameters selection procedure.

\subsection{Experimental set-up}\label{sec:exp_setup}

\paragraph{Anomaly detection models}
In this experimental evaluation, we use two differentiable models with distinct temporal neural networks mechanisms.  The first one, Neural Contextual Anomaly Detection (NCAD) \cite{carmona2021neural}, uses a temporal convolutional network %
and subdivides time series into windows that include a context part. The second one, UnSupervised Anomaly Detection (USAD) \cite{audibert2020}, is based on a LSTM Auto-Encoder and predicts anomalies on suspect windows without explicit context windows. Neither of these models are interpretable-by-design, but both have SOTA performances on the benchmark anomaly detection datasets and reasonable training times (around 90 min). Before evaluating our explainability method, we train these models using the procedure described in their respective papers. More details on these models and their detection performance on the benchmark datasets are reported in Appendix \ref{app:AD_models}.

\paragraph{Datasets}
The four benchmark datasets selected in our experiments are the following:
\begin{itemize}
    \item \textbf{KPI: \footnote{https://github.com/NetManAIOps/KPI-Anomaly-Detection}} this dataset contains 29 univariate time series. It was released in the AIOPS data competition and consists of Key Performance Indicator curves from different internet companies in 1 minute interval.
    \item \textbf{YAHOO: \footnote{https://webscope.sandbox.yahoo.com/catalog.php?datatype=s\&did=70}} this dataset was published by Yahoo labs and consists of 367 real and synthetic univariate time series.
      \item \textbf{Server Machine Dataset (SMD):} this dataset contains 28 time series with 38 dimensions, %
      collected from a machine in large internet companies \cite{su19}.
    \item \textbf{Soil Moisture Active Passive satellite (SMAP):} this NASA dataset published by Hundman et al. (2018) contains 55 times series with 25 dimensions.
\end{itemize}
The main properties of these datasets are given in Table \ref{tab:datasets}. For our evaluation, we use the test sets of each dataset, which correspond to the last 50\% timestamps of each time series \cite{carmona2021neural}. When needed, the training and validation sets contain respectively the first 30\% and subsequent 20\% timestamps. We note that all these datasets have ground-truth anomaly labels on the test set, and in our evaluation, we only compute counterfactual ensemble explanations for the ground-truth anomalies detected by each model (i.e., the \emph{True Positives}). 
Since in practice, our method could be applied on all the detected anomalies, including the \emph{False Positives} (i.e., the observations with anomalous predicted labels that are not ground-truth anomalies), we have also performed a complementary analysis on the latter. The numerical results on the KPI dataset, available in Appendix \ref{app:false_positives}, seem to show that good counterfactual explanations on False Positives are easier to obtain than on True Positives. Consequently, we did not include the former in our numerical evaluation in Section \ref{sec:exp_univariate} since they could eclipse the relative advantages of our method and its variants.

\paragraph{Naive counterfactual ensemble explanation}

As a simple baseline, we propose a procedure that generates a set of counterfactual examples using a basic sampling scheme without requiring any training. The main idea is similar to the Forecasting Set approach, but here the sampling mechanism is "naive". For each tested window $W$ containing an anomaly in $W_S$, we draw a sample by interpolating the anomalous window $W_S$ and a constant window with a random weight. The constant window repeats the observation from the timestamp immediately before the anomaly, i.e., $[W_C]_{L-S}$. Thus, for $i \in [N]$, a sample $\widetilde{W}_S^{naive, i}$ is defined as:
\begin{align}\label{eq:naive_cf}
    \widetilde{W}_S^{naive, i} = w_i W_S + (1 - w_i) X_{-1},
\end{align}
where $w_i
\overset{i.i.d.}{\sim} U[0,1]$ and $X_{-1} = [[W_C]_{L-S}, \dots, [W_C]_{L-S}] \in \mathbb{R}^{S \times D}$. As in Section \ref{sec:forecasting_set}, we also select the samples that are not anomalous under the model, i.e., the naive counterfactual ensemble is finally:
$$I_{N} = \{\widetilde{W}^{naive,i} = [W_C, \widetilde{W}_S^{naive,i}]; \: i \in [N] \: \text{st} \: \forall t \in [S], f(\widetilde{W}^{naive,i})_t < \theta\}$$

\subsection{Explainability metrics}\label{sec:exp_metrics}

To evaluate the utility of our method, we compute the following metrics as proxies of the criteria defined in Section \ref{sec:setup}:
\begin{itemize}
    \item \textbf{Failure rate:} Accounting for Validity or algorithm Correctness, this metric corresponds to the percentage of times our method outputs at least one counterfactual example for the gradient-based methods, and the rejection rate of the samples from the generative scheme in the Forecasting Set approach and naive baseline.
    \item \textbf{Distance:} The Closeness criterion is measured in terms of the Dynamic Time Warping (DTW) distances between each example of the counterfactual ensemble and the original anomalous window. The DTW distance %
    is generally more adapted to time series data than the Euclidean distance.  
    \item \textbf{Implausibility: } since the Plausibility property is not easy to evaluate without expert knowledge of the particular data domain, we decompose it into the three following proxy metrics that cover different notions of deviation from an estimated normal behaviour:
    \begin{itemize}
        \item  DTW distance to a reference time series, here, the median sample from the Forecasting Set approach \textbf{(Implausibility 1)};
        \item Temporal Smoothness (\textbf{Implausibility 2}),  defined as
        \begin{align*}
           \sum_{i=1}^D \sum_{t=1}^{S-1} |[\widetilde{W}_S]_{(t+1)i} - [\widetilde{W}_S]_{ti}|.
        \end{align*}
       
        \item Negative log-likelihood under the probabilistic forecasting distribution $g$, if available \textbf{(Implausibility 3)}.
        \end{itemize}
     We compute the latter metrics for each example of the counterfactual ensemble explanation.
    
    \item \textbf{Diversity: } the range of values spanned in a counterfactual ensemble is evaluated by the variance of the counterfactual examples at each timestamp.
    \item \textbf{Sparsity correctness: } for multivariate time series, if additional information on the anomalous dimensions in the ground-truth anomalies is available, we compute the precision and recall scores of the sparse variants of DPE and ICE in identifying the dimensions to perturb.
\end{itemize}

\subsection{Hyperparameters selection}\label{sec:hp_selection}

The hyperparameters of our counterfactual explanation method with the gradient-based approaches are selected by testing all configurations of $\lambda_1 = \lambda_2, \lambda_T$ in the set $\{0.001, 0.01, 0.1, 1.0\}$, $\sigma_{max}$ in $\{3, 5, 10\}$ and the learning rate of the SGD algorithm in $\{0.01, 0.1, 1.0, 10.0, 1000.0, 10 000.0\}$. As an explainability method can be finely tuned on a particular problem and dataset, the configurations could  be evaluated on all the anomalies in the test set. However, for computational time efficiency reasons, we run this evaluation on 100 randomly chosen anomalies, then evaluate the final performance of the chosen configuration on the entire test set. An exception holds for the the SMAP dataset, which contains less than 100 anomalies detected by the models, therefore we run the configurations' evaluation on the whole test set. For each dataset and detection model, we select the set of hyperparameters having the minimal Implausibility 2, given that the failure rate is kept under a pre-defined level, %
(see Figures \ref{fig:hps_ncad_kpi}, \ref{fig:hps_ncad_yahoo}, \ref{fig:hps_ncad_smap} and \ref{fig:hps_ncad_smd} and tables in Appendix \ref{app:hps}). 
Moreover, we run the SGD algorithm for 1000 iterations and select a maximum of $N = 100$ counterfactual examples along the optimization path.
The hyperparameters of the probabilistic models in the Forecasting Set approach
are reported in Table \ref{tab:hp_forecast} in Appendix \ref{app:hps}. 
Finally, in order to provide a ready-to-use method, we also suggest a default set of hyperparameters in Table \ref{tab:hp_default} in Appendix \ref{app:hps}. For all datasets, models and approaches, we use suspect windows of $S = 10$ timestamps and margin parameter $c=0$.

\subsection{Qualitative analysis}\label{sec:qualitative}

Similarly to image classification settings \cite{zeiler2013visualizing}, visualizations in the time series domain can be human-friendly tools to communicate model explanations, in particular in univariate or low-dimensional settings. %
In our context, we propose to visualize our counterfactual ensemble explanation on the original observations for which a prediction was made, possibly with an added context window (see Section \ref{sec:setup}) and a restricted number of channels. Since the anomaly prediction score is a scalar, we can add the score of each counterfactual example in their representation using an appropriate  color scale. The explanation's recipient can therefore observe how the model's prediction score changes in the counterfactual ensemble, guess  the shape of the model's local decision boundary, and evaluate the range of values spanned by the examples. %

\begin{figure}[H]
    \centering
    \includegraphics[width=0.33\linewidth, trim={1cm 1cm 1cm 1cm},clip]{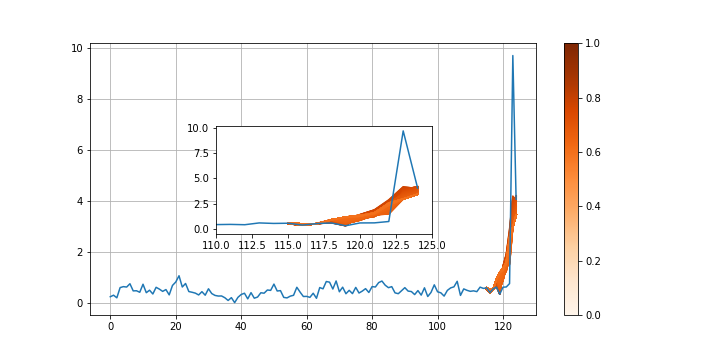}
    \includegraphics[width=0.33\linewidth, trim={1cm 1cm 1cm 1cm},clip]{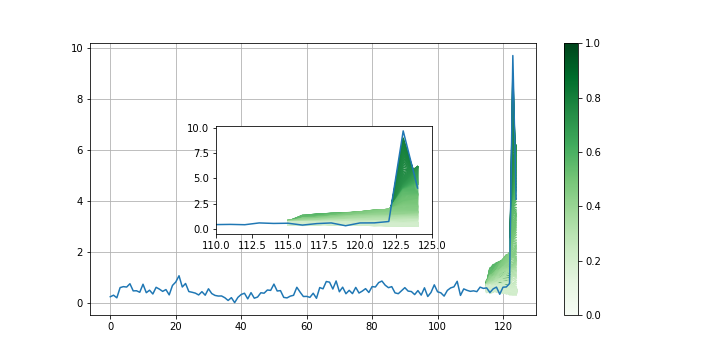}
    \includegraphics[width=0.33\linewidth, trim={1cm 1cm 1cm 1cm},clip]{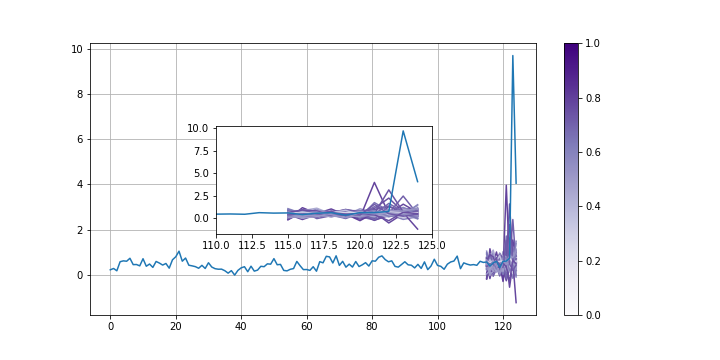}
    \includegraphics[width=0.33\linewidth, trim={1cm 1cm 1cm 1cm},clip]{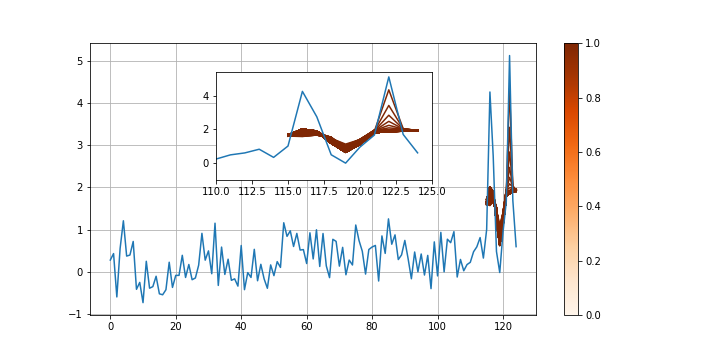}
    \includegraphics[width=0.33\linewidth, trim={1cm 1cm 1cm 1cm},clip]{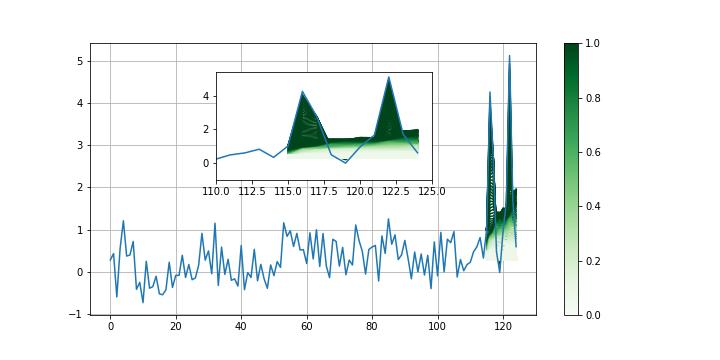}
    \includegraphics[width=0.33\linewidth, trim={1cm 1cm 1cm 1cm},clip]{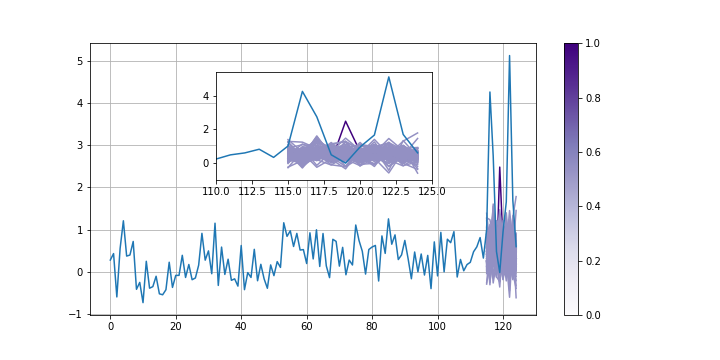}
    \caption{Time series windows contains an anomaly and our counterfactual ensemble explanations, obtained with DPE (left column), ICE (middle column) and FS (right column) from the KPI dataset. The first (resp. second) row corresponds to an anomaly that has been detected by NCAD (resp. USAD). Each window includes a context part of 115 timestamps and an abnormal part of 10 timestamps at the end of the window. The original observations are plotted in blue, while the counterfactual examples appear in red, green or purple color scales for respectively DPE, ICE and FS.}
    \label{fig:comp_kpi}
\end{figure}

On Figure \ref{fig:comp_kpi}, we present a visualization of our method on anomalies from the KPI dataset, detected by NCAD and USAD. We observe that the counterfactual ensemble explanation from DPE (in red color scale), ICE (in green), and FS (in purple) are quite dissimilar, although they all modify only the spike outliers' features and globally lessen their amplitude. In fact, on the one hand, DPE produces counterfactual ensembles that are less diverse than the other approaches, and relatively close to the original input. On the other hand, ICE's counterfactual sets cover a much larger range of values and therefore allows to visualize more clearly how the anomaly score evolves for different magnitudes of the spikes. In contrast, the  counterfactual ensembles generated by FS do not have the aforementioned interpretation but seem to visually correspond to the expected behaviour given the shape of the context windows. Note that additional visualizations of our explanations can be found in Appendix \ref{app:visu_kpi}.

In summary, our counterfactual ensemble explanations effectively contain diverse perturbations of the input time series that change the detected label of the anomalous subsequence, with a small number of altered features. The three approaches, ICE, DPE and FS, bring different insights on the model's prediction, the time series distribution and the possible perturbations to apply to change the former. Their relative advantages may therefore depend on the particular time series context and usage of the counterfactual explanation.

\subsection{Numerical evaluation}\label{sec:exp_univariate}

\begin{table}
    \centering
\begin{tabularx}{\textwidth}{ 
  >{\centering\arraybackslash}X 
  >{\centering\arraybackslash}X 
  >{\centering\arraybackslash}X 
  >{\centering\arraybackslash}X 
  >{\centering\arraybackslash}X 
  >{\centering\arraybackslash}X 
  >{\centering\arraybackslash}X %
  }
  \hline
{Dataset} & Dimensions & Number of time series & Total number of timestamps & Total number of anomalies in test set  \\
\hline
KPI & 1 & 29 & 5922913 & 54560 \\
Yahoo & 1 & 367 & 609666 & 2963  \\
SMD & 38 & 28 & 1416825 & 29444  \\
SMAP & 25 & 55 & 584860 & 57079 \\
\end{tabularx}
    \caption{Succinct description of the four benchmark datasets}
    \label{tab:datasets}
\end{table}

The numerical results discussed in this section are obtained in the set-up described in Section \ref{sec:exp_setup}. %
However, due to the space limitation, some of these results have been moved to Appendix \ref{app:add_results}. We also add a partial sensitivity analysis of our method in Appendix \ref{app:sensitivity}.

The results on univariate datasets (see Table \ref{tab:res_ncad_yahoo} and Table \ref{tab:res_ncad_kpi} in Appendix \ref{app:res_kpi}), %
show that our method has fairly small failure rates (except for the Yahoo dataset and the USAD model). In particular a rate smaller than 10\% can be achieved with at least one variant in most pairs (model, dataset), leading to a consequent improvement over the naive procedure. We note that while the DPE variant seems to be valid more often than ICE on the NCAD model, it is the contrary for the USAD model; this difference is possibly due to the distinct internal mechanisms of these models. %

Moreover, the analysis of the other explainability metrics supports the qualititative interpretation from Section \ref{sec:qualitative}. The Distance metric confirms than the gradient-based approaches, DPE and ICE, provides in almost all cases the closest counterfactuals in average, i.e. the least perturbed examples. Note that it sometimes occur that the naive baseline has a small distance, however its high failure rate probably indicates it might be a side-effect of the way examples are picked.
Moreover, the Implausibility metrics validate the observation that FS generates the most realistic counterfactual examples in average, in particular in terms of Implausibility 1 (distance to median forecast sample) and Implausibility 3 (NLL under the probabilistic forecasting distribution). This is in fact quite expected since these quantities are directly derived from the forecasting sampling scheme. However, these counterfactuals are less smooth (higher score in Implausibility 2) than for DPE and ICE, which regularize the time series smoothness in the objective functions \eqref{eq:objective_dynamap} and \eqref{eq:objective_icod}.

Finally, DPE and ICE provide a more diverse counterfactual ensemble in most cases in general, but their relative ranking is not clear from these experiments. We conjecture that this metric is particularly sensitive to the learning rate of the SGD algorithm, and the subsampling procedure after the objective minimization (see Section \ref{sec:gradient_methods}). In Appendix \ref{app:sensitivity}, we test our first hypothesis on a small sample of anomalies. We observe in this case that the Diversity criterion is consistently higher for ICE, and greatly increases with the learning rate, at the cost of a higher failure rate.

\begin{table}
    \centering
  \begin{tabularx}{\textwidth}{ 
   >{\centering\arraybackslash}X 
   >{\centering\arraybackslash}X 
   >{\centering\arraybackslash}X 
   >{\centering\arraybackslash}X 
   >{\centering\arraybackslash}X 
   >{\centering\arraybackslash}X 
   >{\centering\arraybackslash}X  }
          \hline
            & \multicolumn{6}{c}{\textbf{NCAD} on Yahoo}\\
  \hline
          Method & Failures (\%) & Distance  & Implausibility 1  & Implausibility 2  & Implausibility 3 & Diversity \\
         \hline
DPE  &          \textbf{9.2} &   2.49 (4.91) &                   1.23 (1.37) &                 1.42 (2.19) &          2.21 (4.76) &      0.01 \\
ICE    &         17.4 &   \textbf{1.54 (1.21)} &                   0.78 (1.37) &                 2.26 (1.67) &          1.40 (5.17) &      0.05 \\
FS &        56.6 &  6.06 (16.38) &                   \textbf{0.27 (0.22)} &                 3.36 (1.99) &         \textbf{-0.29 (0.79)} &      \textbf{0.10} \\
Naive    &         72.2 &   2.69 (5.29) &                   1.04 (1.26) &                 \textbf{1.32 (1.74)} &          1.89 (3.34) &      0.05 \\
\hline
\hline
    \end{tabularx}
  \begin{tabularx}{\textwidth}{ 
   >{\centering\arraybackslash}X 
   >{\centering\arraybackslash}X 
   >{\centering\arraybackslash}X 
   >{\centering\arraybackslash}X 
   >{\centering\arraybackslash}X 
   >{\centering\arraybackslash}X 
   >{\centering\arraybackslash}X  }
          \hline
                      & \multicolumn{6}{c}{\textbf{USAD} on Yahoo}\\
  \hline
          Method & Failures (\%) & Distance  & Implausibility 1  & Implausibility 2  & Implausibility 3 & Diversity \\
         \hline
DPE  &          29.1 &   5.20 (18.00) &                  6.42 (26.81) &                 0.42 (2.00) &          3.74 (6.96) &      0.05 \\
ICE     &         \textbf{25.5} &   6.66 (25.54) &                  2.68 (11.46) &                 \textbf{0.40 (1.16)} &          2.48 (4.61) &      \textbf{3.23} \\
FS &          65.1 &  14.48 (46.03) &                  \textbf{ 0.48 (0.72)} &                 0.55 (0.58) &         \textbf{-0.11 (1.12)} &      0.61 \\
Naive    &         45.8 &   \textbf{4.82 (18.36)} &                  2.85 (16.31) &                 0.52 (1.60) &          3.25 (6.00) &      3.19 \\
         \hline
    \end{tabularx}
    \caption{Performance of our explainability method and the naive baseline in terms of Validity, Closeness, Plausibility and Diversity on the Yahoo dataset and the NCAD (first panel) and USAD (second panel) anomaly detection models. We report the average scores and standard deviations (in brackets) over the counterfactual ensembles. We recall that \emph{Implausibility 1} is the DTW distance to the median forecasting sample, \emph{Implausibility 2} is the temporal smoothness, and \emph{Implausibility 3} is the negative log-likelihood under the probabilistic forecasting output distribution. For all metrics except \emph{Diversity}, we assume that a lower value is better, and the best score is highlighted in bold.}
    \label{tab:res_ncad_yahoo}
\end{table}

The multivariate experiments (see Table \ref{tab:res_ncad_smd} and Table \ref{tab:res_ncad_smap} in Appendix \ref{app:res_kpi}) showcase that our method also generates valid counterfactual ensemble explanations in this more challenging setting, with even a failure rate of 0\% for the USAD model. Our method fails more frequently on the NCAD model, however, the sparse variants are more often successful. This seems to show that imposing a sparsity constraint over the modified dimensions also helps to find valid counterfactuals. Consistently with the univariate datasets, FS produces the most realistic counterfactual examples while the gradient-based approach achieves a better Distance score. We note that in this case the Implausibility 3 metric is not available since the forecast distribution likelihood function in the DeepVAR model is not available \footnote{https://ts.gluon.ai/api/gluonts/gluonts.model.deepvar.html}. Moreover, the sparse variants seem to correctly identify some of the anomalous channels (precision greater than 0.6 for the USAD model).

Nonetheless, we noted the greater difficulty of tuning the hyperparameters of our method and ranking its variants on these high-dimensional datasets compared to univariate data. In the latter, the default set of hyperparameters achieves an acceptable performance and allows to quickly compare the relative advantages of an approach for a specific pair (detection model, dataset). We therefore conclude by recalling that example-based explainability methods for multivariate time series are still in their early developments, and providing general methods and tuning procedures to generate useful explanations over the instances of a dataset is still an open problem.

\begin{table}
    \centering
    \begin{tabularx}{\textwidth}{ 
  >{\centering\arraybackslash}X 
  >{\centering\arraybackslash}X 
  >{\centering\arraybackslash}X 
  >{\centering\arraybackslash}X 
  >{\centering\arraybackslash}X 
  >{\centering\arraybackslash}X
  >{\centering\arraybackslash}X  } %
          \hline
                      & \multicolumn{5}{c}{\textbf{NCAD} on SMD}\\
  \hline
          Method & Failures (\%) & Precision / Recall  &    Distance & Implausibility 1  & Implausibility 2 & Diversity\\
\hline
DPE        &         \textbf{17.1} &      -   &    \textbf{8.46 (13.07)} &                50.76 (110.40) &               12.12 (28.13) &      1.21 \\
ICE           &         42.9 &      - &  79.62 (120.27) &                 23.69 (28.24) &               47.73 (58.32) &      \textbf{4639.11} \\
Sparse DPE &         20.0 &      \textbf{0.22} /   0.10  &   36.12 (77.87) &                 29.03 (54.36) &                5.61 (11.30) &   \textbf{4687.05}\\
Sparse ICE    &         20.0 &      0.20 /   \textbf{0.33}  &   26.01 (37.07) &                62.65 (107.25) &               10.88 (16.72) &    174.39 \\
FS       &         30.0 &     -  &  78.46 (157.57) &                   \textbf{1.62 (2.33)} &                 \textbf{1.49 (1.31)}  &     35.88\\
Naive          &         79.8 &         -  &   25.06 (53.66) &                 45.45 (93.03) &                9.42 (18.79) &   3255.92 \\

         \hline
    \end{tabularx}
    \begin{tabularx}{\textwidth}{ 
  >{\centering\arraybackslash}X 
  >{\centering\arraybackslash}X 
  >{\centering\arraybackslash}X 
  >{\centering\arraybackslash}X 
  >{\centering\arraybackslash}X 
  >{\centering\arraybackslash}X
  >{\centering\arraybackslash}X  } %
          \hline
                                & \multicolumn{5}{c}{\textbf{USAD} on SMD}\\
  \hline
          Method & Failures (\%) & Precision / Recall  &    Distance & Implausibility 1  & Implausibility 2 & Diversity \\
\hline
DPE        &          \textbf{0.0} &     -  &  139.02 (261.44) & 258.31 (464.18) &               41.19 (79.11) &  \textbf{23339.20} \\
ICE           &          \textbf{0.0} &        -  &     \textbf{31.81 (9.27)} &               342.19 (708.88) &               22.64 (45.16) &      0.52 \\
Sparse DPE &          \textbf{0.0} &      \textbf{0.68} /  0.07  &  115.48 (206.46) &               293.70 (679.88) &               19.84 (34.98) &    105.45\\
Sparse ICE    &          \textbf{0.0} &      0.61 /   \textbf{0.28} &  216.44 (316.05) &               172.58 (475.15) &                \textbf{8.35 (17.52)}  &    477.43 \\
FS       &          \textbf{0.0} &     -  &  366.57 (672.44) &                 \textbf{18.10 (48.25)} &                \textbf{8.57 (20.63)} &     12175.65 \\
Naive          &         73.4 &     -   &    49.83 (60.13) &               475.42 (879.38) &               27.45 (45.43) &    649.44 \\
         \hline
    \end{tabularx}
    \caption{Performance of our explainability method and the naive baseline in terms of Validity, Closeness, Plausibility and Diversity on the SMD dataset and the NCAD (first panel) and USAD (second panel) anomaly detection models. We report the average scores and standard deviations (in brackets) over the counterfactual ensemble. We recall that \textit{Implausibility 1} is the DTW distance to the median forecasting sample and \textit{Implausibility 2} is the temporal smoothness. For all metrics except \emph{Diversity}, \emph{Precision} and \emph{Recall}, we assume that a lower value is better, and the best score is highlighted in bold.}
    \label{tab:res_ncad_smd}
\end{table}

\section{Concluding remarks}\label{sec:discussion}

This work proposed a novel method for generating explanations for time series anomalies and detection models. Our real-world experiments show that the counterfactual framework, augmented with an ensemble approach, improves the interpretability of time series anomaly detection models, and can help their users identify the possible actions to take in consequence. Since there is generally little \textit{a priori} knowledge on the possible anomalies, we compared several approaches that leverage domain-specific perturbations and anomaly sparsity in high-dimensional settings. Additionally, we have proposed a gradient-free approach that uses probabilistic forecasting techniques as a generative scheme. Although our model-agnostic method offers greater flexibility, better explainability performances might be achieved if more assumptions are put on the detection model. In particular, similarly to \cite{rodriguez2021trivial}, we could adapt our gradient-based approach to use the internal representations of the model rather than the raw time series.


\printbibliography

\appendix

\section{Technical details and performance of the selected anomaly detection models}\label{app:AD_models}

In this section, we provide some technical details on the two anomaly detection models selected for the evaluation of our explainability method reported in Section \ref{sec:experiments}. In Table \ref{tab:ad_performances}, we report their anomaly detection performances on the benchmark datasets, after training with the hyperparameters' sets reported in their respective papers %
when  available. Otherwise, we select the models' hyperparameters on a validation set (20\% of the time series) using the best adjusted F1-score. 

\paragraph{ Neural Contextual Anomaly Detection (NCAD) \cite{carmona2021neural} :} This method splits time series into subwindows $(W^i)_i$ and embeds them using a temporal convolutional network (TCN). Each $W^i$ is subdivided into a context part and a suspect part (typically much smaller than the former), i.e., $W^i = [W_C^i, W_S^i]$. An embedding of the context window $W_C^i$ is also computed by the TCN, then the distance between the embeddings of $W_i$, denoted $z^i$, and $W_C^i$, denoted $z_C^i$, is evaluated. The algorithm finally labels $W_S^i$ as anomalous if the latter distance is greater than a chosen threshold, i.e, if $d(z^i,z^i_C) > \eta$ with $d(.,.)$ the Euclidean distance for instance and $\eta > 0$. The intuition behind this method is that a large distance between the embeddings of a window and its context part means that the suspect part induces a significant shift of $z_C^i$ in the embedding space. Since the embedding of the context window should reflect the normal behaviour, this deviation thus indicates the presence of an anomaly in $W_S^i$. For our experiments, we use the open-source implementation. \footnote{
\url{https://github.com/Francois-Aubet/gluon-ts/tree/adding_ncad_to_nursery/src/gluonts/nursery/ncad}}

\paragraph{UnSupervised Anomaly Detection (USAD) \cite{audibert2020}:} This reconstruction model splits time series into subwindows that are reconstructed by a LSTM-based AutoEncoder. The latter contains a neural network, called encoder, that embeds each window into a latent representation, and another neural network, called decoder, that maps back the embedding into the original input space. The reconstruction error, i.e., the distance in the time series domain between the original input and the reconstructed output, is used as an anomaly score (a high value of this error leads to the corresponding window to be labelled as anomalous). 
We use the open source implementation provided by the authors \footnote{\url{https://curiousily.com/posts/time-series-anomaly-detection-using-lstm-autoencoder-with-pytorch-in-python/}} and the hyperparameters provided in the paper for the two multivariate data sets, i.e. SMD and SMAP. For the KPI dataset, the final USAD model is trained for 80 epochs and has windows of size 5, hidden size of 10 and downsampling rate of 0.01. For the Yahoo data, the window size is 10, hidden size of 10 and downsampling rate of 0.05.

\begin{table}[H]
    \centering
    \begin{tabularx}{\textwidth}{ 
   >{\centering\arraybackslash}X 
   >{\centering\arraybackslash}X 
   >{\centering\arraybackslash}X 
   >{\centering\arraybackslash}X 
   >{\centering\arraybackslash}X  } %
          \hline
          Model & KPI & Yahoo & SMD & SMAP \\
         \hline
NCAD &   0.789 & 0.772 & 0.806 & 0.922 \\
USAD &   0.946 & 0.741 & 0.643 & 0.972 \\
         \hline
    \end{tabularx}
    \caption{F1-scores of the two anomaly detection models, i.e., NCAD and USAD, on the four benchmark datasets. }
    \label{tab:ad_performances}
\end{table}

\section{Additional numerical results}\label{app:add_results}

In this section, we report quantitative evaluations of our explainability method that could not be included in the main text due to space limitation. This section notably contains the results on two benchmark datasets using the procedure described in Section \ref{sec:experiments}, and an additional analysis on False Positives.

\subsection{Numerical evaluation on the KPI and SMAP datasets}\label{app:res_kpi}

The results on the KPI and SMAP dataset are respectively in Table \ref{tab:res_ncad_kpi} and Table \ref{tab:res_ncad_smap}. Note that these results are included in the discussion in Section \ref{sec:exp_univariate}. 

\begin{table}
    \centering
\begin{tabularx}{\textwidth}{ 
   >{\centering\arraybackslash}X 
   >{\centering\arraybackslash}X 
   >{\centering\arraybackslash}X 
   >{\centering\arraybackslash}X 
   >{\centering\arraybackslash}X 
   >{\centering\arraybackslash}X 
   >{\centering\arraybackslash}X  }
  \hline
  & \multicolumn{6}{c}{\textbf{NCAD} on KPI}\\
  \hline
{Method} & Failures (\%) &       Distance & Implausibility 1 & Implausibility 2  & Implausibility 3 & Diversity \\
\hline
DPE  &           \textbf{3.9} &    5.94 (15.78) &                   2.16 (4.71) &                3.21 (29.62) &          2.74 (2.57) &     \textbf{ 1.18} \\
ICE     &         19.6 &     \textbf{3.08 (1.21)} &                15.31 (115.12) &              31.67 (206.70) &          2.07 (2.06) &      0.26 \\
FS &          6.0 &  32.05 (173.76) &                   \textbf{0.21 (0.20)} &                 \textbf{2.42 (1.97)} &         \textbf{-0.56 (1.14)} &      0.12 \\
Naive    &         53.4 &   11.82 (74.03) &                   2.90 (4.49) &                 4.57 (7.64) &          3.48 (2.51) &      0.54 \\
\hline
\end{tabularx}
\begin{tabularx}{\textwidth}{ 
   >{\centering\arraybackslash}X 
   >{\centering\arraybackslash}X 
   >{\centering\arraybackslash}X 
   >{\centering\arraybackslash}X 
   >{\centering\arraybackslash}X 
   >{\centering\arraybackslash}X 
   >{\centering\arraybackslash}X  }
  \hline
    & \multicolumn{6}{c}{\textbf{USAD} on KPI}\\
  \hline
      Method & Failures (\%) & Distance  & Implausibility 1  & Implausibility 2  & Implausibility 3 & Diversity \\
\hline
DPE  &           5.0 &  25.22 (121.60) &                  9.40 (65.02) &                 1.03 (8.16) &          3.13 (3.47) &     13.10 \\
ICE     &          \textbf{3.5} &     \textbf{6.52 (7.30)} &                  4.99 (63.31) &                 0.50 (4.43) &          1.27 (1.98) &      0.28 \\
FS &          6.8 &  38.56 (189.88) &                  \textbf{ 0.33 (0.29)} &                 \textbf{0.38 (0.28)} &         \textbf{-0.08 (1.12)} &      0.26 \\
Naive    &         45.4 &  31.93 (154.62) &                   2.77 (3.93) &                 1.42 (6.01) &          2.81 (2.48) &     \textbf{69.88} \\
\hline
\end{tabularx}
    \caption{Performance of our explainability method and the naive baseline in terms of Validity, Closeness, Plausibility and Diversity on the KPI dataset and the NCAD (first panel) and USAD (second panel) anomaly detection models. We report the average scores and standard deviations (in brackets) over the counterfactual ensemble. We recall that \emph{Implausibility 1} is the DTW distance to the median forecasting sample, \emph{Implausibility 2} is the temporal smoothness, and \emph{Implausibility 3} is the negative log-likelihood under the probabilistic forecasting output distribution. For all metrics except \emph{Diversity}, we assume that a lower value is better, and the best score is highlighted in bold.}
    \label{tab:res_ncad_kpi}
\end{table}

\begin{table}
    \centering
    \begin{tabularx}{\textwidth}{ 
   >{\centering\arraybackslash}X 
   >{\centering\arraybackslash}X 
   >{\centering\arraybackslash}X 
   >{\centering\arraybackslash}X
   >{\centering\arraybackslash}X
   >{\centering\arraybackslash}X  } %
          \hline
                      & \multicolumn{5}{c}{\textbf{NCAD}}\\
  \hline
          Method & Failures (\%) & Diversity &    Distance & Implausibility 1  & Implausibility 2 \\
\hline
DPE       &         41.7 &      0.002 &  0.19 (0.40) &                   0.21 (0.28) &                 \textbf{0.01 (0.03)} \\
DPE sparse &         27.8 &      0.004 &  0.22 (0.42) &                   0.29 (0.38) &                 0.03 (0.04) \\
ICE          &          \textbf{5.6} &      \textbf{0.067} &  0.26 (0.14) &                   0.39 (0.37) &                 0.15 (0.09) \\
ICE sparse    &         23.6 &      0.016 &  0.15 (0.08) &                   0.22 (0.21) &                 0.09 (0.06) \\
FS      &         87.6 &      0.012 &  0.56 (0.77)  &                 \textbf{0.05 (0.04)} &                   0.05 (0.04)\\
Naive          &         84.5 &      0.003 &  \textbf{0.06 (0.08)} &                   0.09 (0.03) &                 0.02 (0.03) \\
         \hline
    \end{tabularx}
    \begin{tabularx}{\textwidth}{ 
  >{\centering\arraybackslash}X 
   >{\centering\arraybackslash}X 
   >{\centering\arraybackslash}X 
   >{\centering\arraybackslash}X 
   >{\centering\arraybackslash}X
   >{\centering\arraybackslash}X } %
          \hline
                                & \multicolumn{5}{c}{\textbf{USAD}}\\
  \hline
          Method & Failures (\%) & Diversity &    Distance & Implausibility 1  & Implausibility 2 \\
\hline
DPE        &          \textbf{0.0} &      0.02 &  0.62 (0.65) &                   0.96 (0.53) &                 0.06 (0.05) \\
DPE sparse &          \textbf{0.0} &      0.02 &  0.78 (0.85) &                   0.82 (0.50)  &                 0.05 (0.06) \\
ICE           &          \textbf{0.0} &      0.17 &  0.74 (0.75) &                   0.87 (0.43) &                 \textbf{0.04 (0.03)} \\
ICE sparse    &          \textbf{0.0} &      \textbf{0.18} &  0.72 (0.76) &                   0.88 (0.42) &                 0.06 (0.02) \\
FS       &         56.8 &      0.02 &  2.23 (1.14) &                   \textbf{0.09 (0.01)} &                 0.10 (0.03) \\
Naive          &         46.9 &      0.01 &  \textbf{0.14 (0.04)} &                   0.23 (0.02) &                 0.07 (0.02) \\
         \hline
    \end{tabularx}
    \caption{Performance of our explainability method and the naive baseline in terms of Validity, Closeness, Plausibility and Diversity on the SMAP dataset and the NCAD (first panel) and USAD (second panel) anomaly detection models. We report the average scores and standard deviations (in brackets) over the counterfactual ensemble. We recall that \textit{Implausibility 1} is the DTW distance to the median forecasting sample and \textit{Implausibility 2} is the temporal smoothness. For all metrics except \emph{Diversity}, \emph{Precision} and \emph{Recall}, we assume that a lower value is better, and the best score is highlighted in bold.}
    \label{tab:res_ncad_smap}
\end{table}

\subsection{Numerical evaluation on False Positives}\label{app:false_positives}

In the practical use of anomaly detection models, explanations can also be needed when the model wrongly detects an anomaly a time series. We recall that we call False Positives the anomalies detected by the model that are not ground-truth anomalies. We present here a numerical evaluation on the False Positives detected by NCAD in the KPI benchmarck dataset. The results in Table \ref{tab:res_ncad_kpi_fp} can be compared to the results obtained on True Positives (i.e., the ground-truth, detected anomalies) reported in the first panel of Table \ref{tab:res_ncad_kpi} . We observe that in this case ICE  achieves 0\% failure rate (instead of almost 20 \%), and the naive method has also a significantly smaller number of failures. Moreover, all methods seem to perform better in terms of the Distance and Implausibility metrics. This is probably due to the fact that False Positives need less perturbation to become not anomalous for the model, e.g. if they lie close to the model's local decision boundary. Therefore they may inherently be less distant to the normal behaviour than True Positives and thus easier instances for our counterfactual explanation method. Besides, the Diversity metric is smaller for DPE and ICE, likely as another effect of the smaller amount of perturbation needed.

\begin{table}[H]
    \centering
\begin{tabularx}{\textwidth}{ 
   >{\centering\arraybackslash}X 
   >{\centering\arraybackslash}X 
   >{\centering\arraybackslash}X 
   >{\centering\arraybackslash}X 
   >{\centering\arraybackslash}X 
   >{\centering\arraybackslash}X 
   >{\centering\arraybackslash}X  }
  \hline
    & \multicolumn{6}{c}{\textbf{NCAD}}\\
  \hline
      Method & Failures (\%) & Distance  & Implausibility 1  & Implausibility 2  & Implausibility 3 & Diversity \\
\hline
DPE  &          8.8 &  \textbf{2.22 (1.87)} &                   2.44 (2.50) &                 2.36 (2.05) &          2.15 (2.22) &      0.02 \\
ICE     &          \textbf{0.0} &  4.36 (3.00) &                   \textbf{0.28 (0.45)} &                 \textbf{0.61 (0.34)} &          0.22 (0.93) &      0.12 \\
FS &          6.6 &  4.17 (2.91) &                   0.30 (0.31) &                 3.54 (3.61) &         \textbf{-0.22 (0.95)} &      \textbf{0.43} \\
Naive    &         33.3 &  2.74 (2.22) &                   2.19 (2.43) &                 2.97 (2.56) &          2.79 (2.29) &      0.16 \\

\hline
\end{tabularx}
    \caption{Performance of our explainability method and the naive baseline in terms of Validity, Closeness, Plausibility and Diversity on the false positives in the KPI data detected by the NCAD model. We report the average scores and standard deviations (in brackets) over the counterfactual ensemble. We recall that \emph{Implausibility 1} is the DTW distance to the median forecasting sample, \emph{Implausibility 2} is the temporal smoothness, and \emph{Implausibility 3} is the negative log-likelihood under the probabilistic forecasting output distribution. For all metrics except \emph{Diversity}, we assume that a lower value is better, and the best score is highlighted in bold.}
    \label{tab:res_ncad_kpi_fp}
\end{table}

\section{Complementary visualizations of the explanations}\label{app:visu_kpi}

\begin{figure}
    \centering
       \includegraphics[width=0.33\linewidth, trim={1cm 1cm 1cm 1cm},clip]{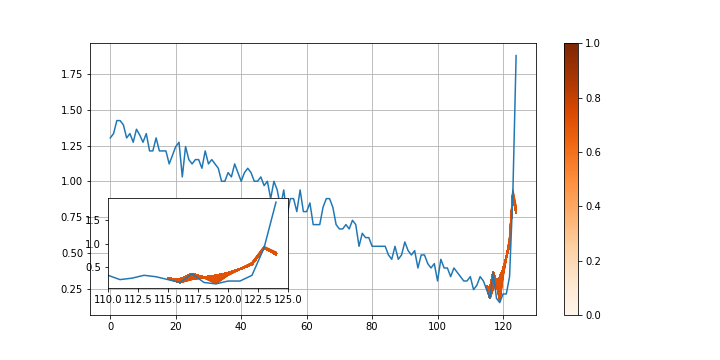}
    \includegraphics[width=0.33\linewidth, trim={1cm 1cm 1cm 1cm},clip]{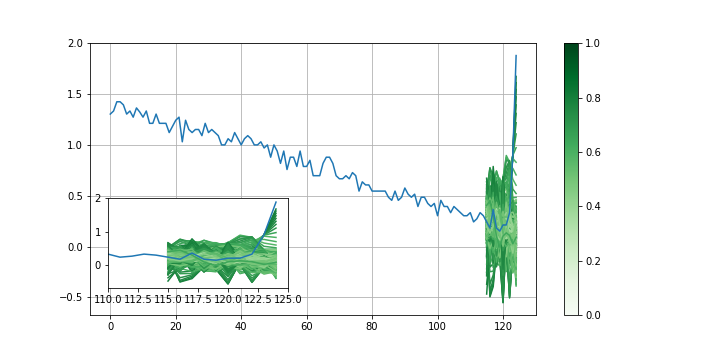}
    \includegraphics[width=0.33\linewidth, trim={1cm 1cm 1cm 1cm},clip]{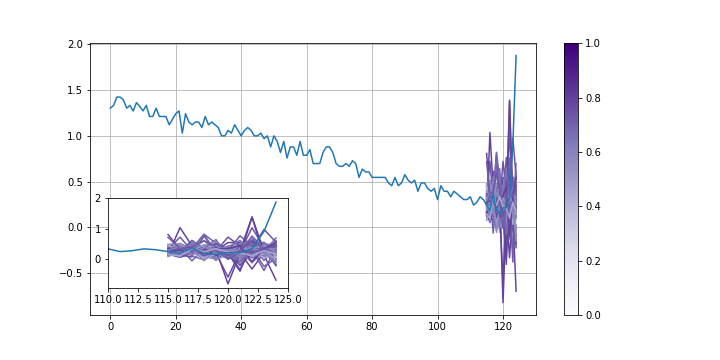}
        \includegraphics[width=0.33\linewidth, trim={1cm 1cm 1cm 1cm},clip]{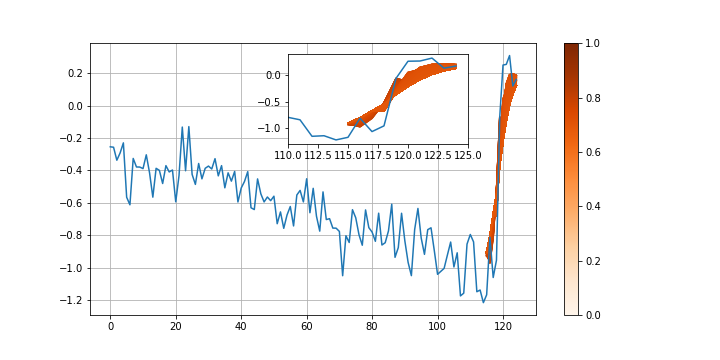}
    \includegraphics[width=0.33\linewidth, trim={1cm 1cm 1cm 1cm},clip]{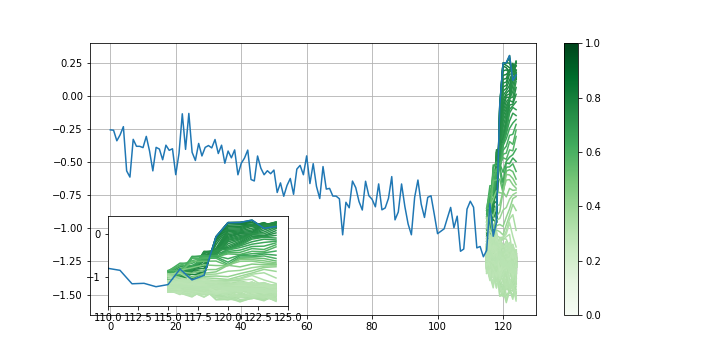}
    \includegraphics[width=0.33\linewidth, trim={1cm 1cm 1cm 1cm},clip]{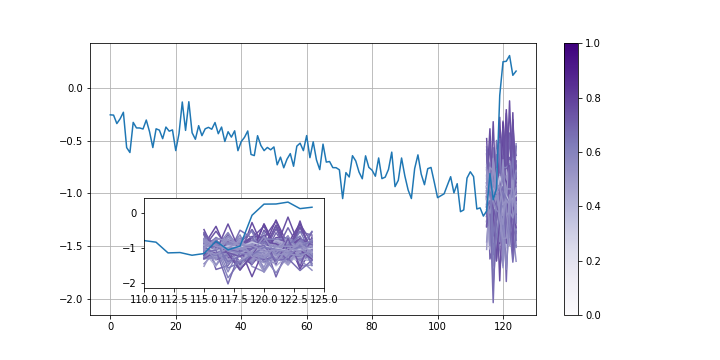}
    \includegraphics[width=0.33\linewidth, trim={1cm 1cm 1cm 1cm},clip]{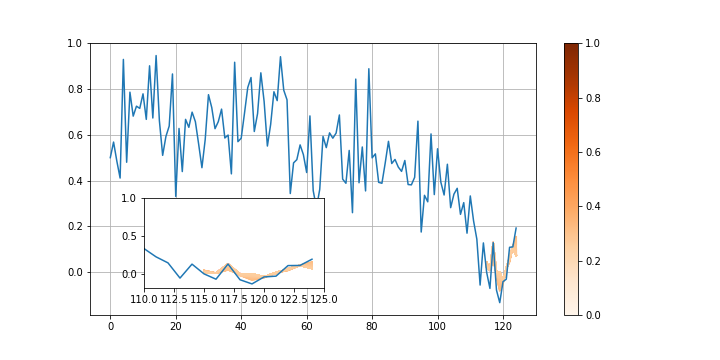}
    \includegraphics[width=0.33\linewidth, trim={1cm 1cm 1cm 1cm},clip]{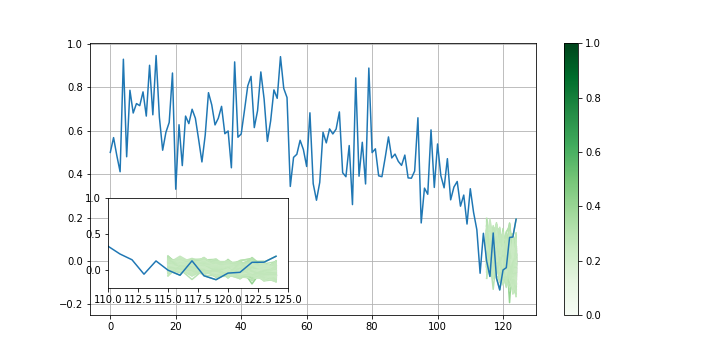}
    \includegraphics[width=0.33\linewidth, trim={1cm 1cm 1cm 1cm},clip]{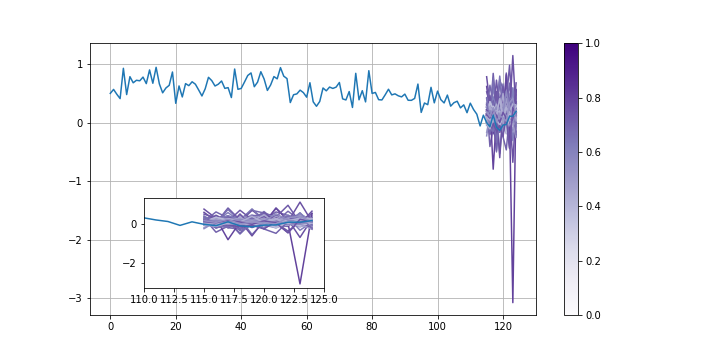}
    \includegraphics[width=0.33\linewidth, trim={1cm 1cm 1cm 1cm},clip]{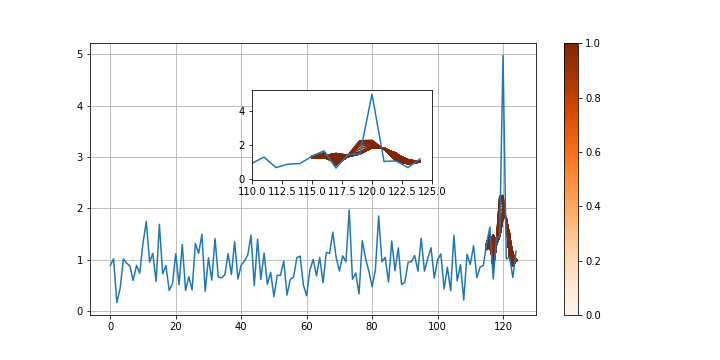}
    \includegraphics[width=0.33\linewidth, trim={1cm 1cm 1cm 1cm},clip]{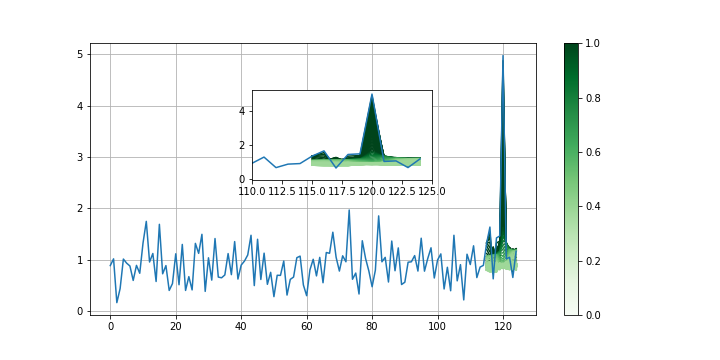}
\includegraphics[width=0.33\linewidth, trim={1cm 1cm 1cm 1cm},clip]{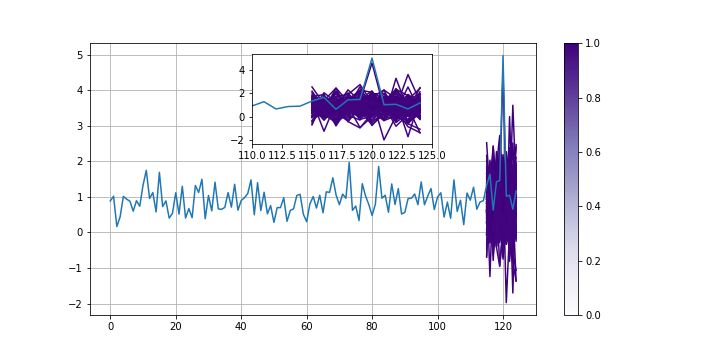}
\includegraphics[width=0.33\linewidth, trim={1cm 1cm 1cm 1cm},clip]{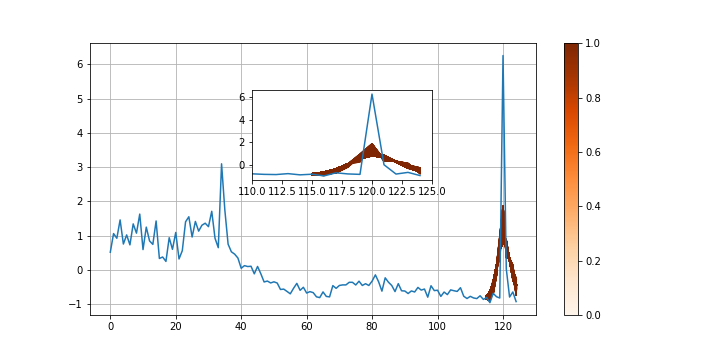}
\includegraphics[width=0.33\linewidth, trim={1cm 1cm 1cm 1cm},clip]{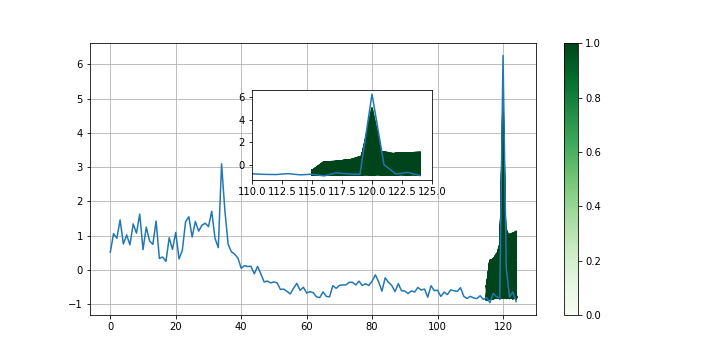}
\includegraphics[width=0.33\linewidth, trim={1cm 1cm 1cm 1cm},clip]{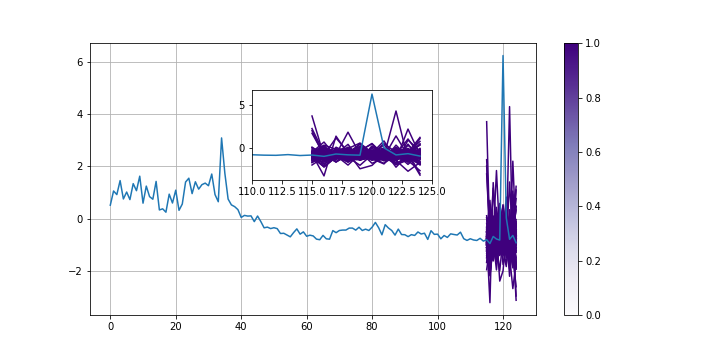}
    \caption{Anomalous windows and counterfactual ensemble explanations obtained with DPE (left column), ICE (middle column) and FS (right column) on anomalies in the KPI and Yahoo datasets detected by the NCAD model. The rows correspond to different anomalies. The windows include a context part of 115 timestamps and an abnormal part of 10 timestamps. The original subsequence is plotted in blue, while the explanations are in red, green or purple colors for the different variants. }
    \label{fig:vis_kpi_ncad}
\end{figure}

\begin{figure}
    \centering
        \includegraphics[width=0.33\linewidth, trim={1cm 1cm 1cm 1cm},clip]{figures/kpi_usad_cfs_ts_4_anomaly_10_dynamap.png}
    \includegraphics[width=0.33\linewidth, trim={1cm 1cm 1cm 1cm},clip]{figures/kpi_usad_cfs_ts_4_anomaly_10_icod.png}
    \includegraphics[width=0.33\linewidth, trim={1cm 1cm 1cm 1cm},clip]{figures/kpi_usad_cfs_ts_4_anomaly_10_forecast.png}
    \includegraphics[width=0.33\linewidth, trim={1cm 1cm 1cm 1cm},clip]{figures/kpi_usad_cfs_ts_4_anomaly_60_dynamap.png}
    \includegraphics[width=0.33\linewidth, trim={1cm 1cm 1cm 1cm},clip]{figures/kpi_usad_cfs_ts_4_anomaly_60_icod.png}
    \includegraphics[width=0.33\linewidth, trim={1cm 1cm 1cm 1cm},clip]{figures/kpi_usad_cfs_ts_4_anomaly_60_forecast.png}

           \includegraphics[width=0.33\linewidth, trim={1cm 1cm 1cm 1cm},clip]{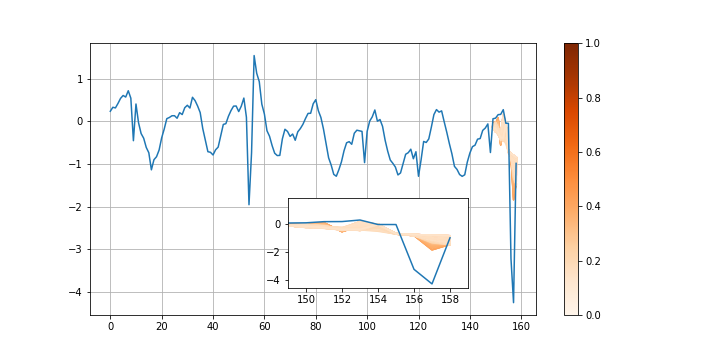}
    \includegraphics[width=0.33\linewidth, trim={1cm 1cm 1cm 1cm},clip]{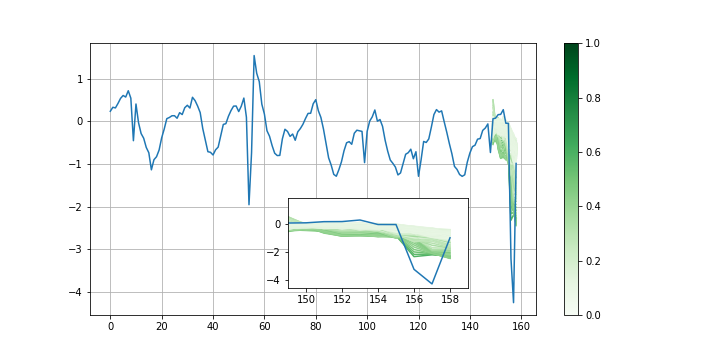}
    \includegraphics[width=0.33\linewidth, trim={1cm 1cm 1cm 1cm},clip]{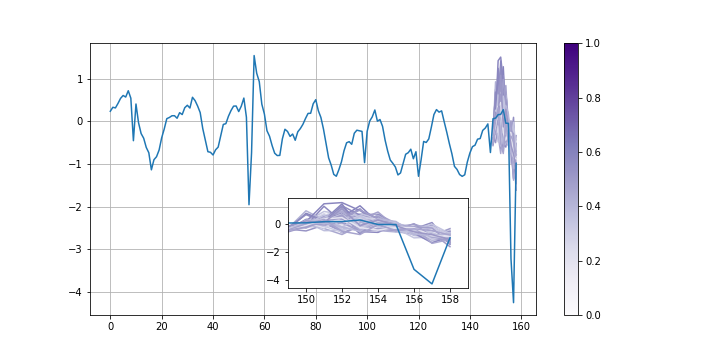}
        \includegraphics[width=0.33\linewidth, trim={1cm 1cm 1cm 1cm},clip]{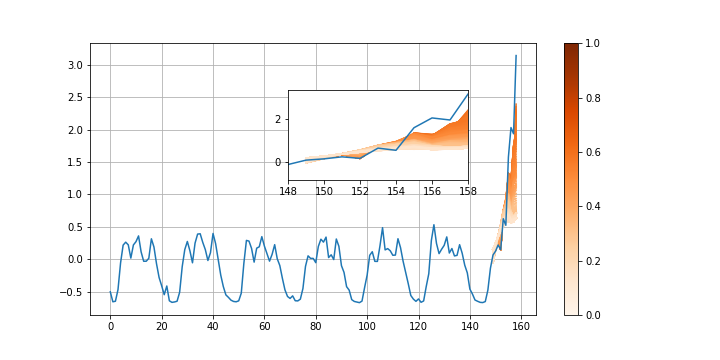}
    \includegraphics[width=0.33\linewidth, trim={1cm 1cm 1cm 1cm},clip]{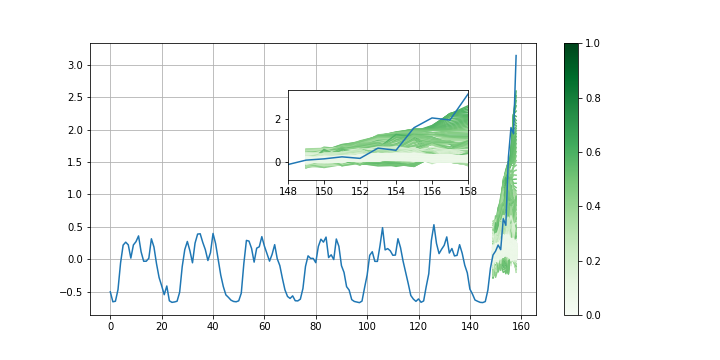}
    \includegraphics[width=0.33\linewidth, trim={1cm 1cm 1cm 1cm},clip]{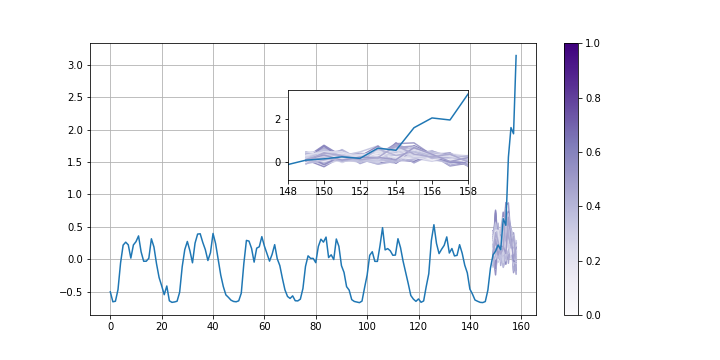}
           \includegraphics[width=0.33\linewidth, trim={1cm 1cm 1cm 1cm},clip]{figures/yahoo_ncad_10_ts_19_anomaly_0_dynamap.png}
    \includegraphics[width=0.33\linewidth, trim={1cm 1cm 1cm 1cm},clip]{figures/yahoo_ncad_10_ts_19_anomaly_0_icod.png}
    \includegraphics[width=0.33\linewidth, trim={1cm 1cm 1cm 1cm},clip]{figures/yahoo_ncad_10_ts_19_anomaly_0_forecast.png}

        \caption{Anomalous windows and counterfactual ensemble explanations obtained with DPE (left column), ICE (middle column) and FS (right column) on anomalies in the KPI and Yahoo datasets detected by the USAD model. The rows correspond to different anomalies. The windows include a context part of 115 timestamps and an abnormal part of 10 timestamps. The original subsequence is plotted in blue, while the explanations are in red, green or purple colors for the different variants.}
    \label{fig:vis_kpi_usad}
\end{figure}

\begin{figure}
\begin{subfigure}[b]{.45\textwidth}
     \centering
    \includegraphics[width=1.3\textwidth, trim =4cm 0cm 7cm 0cm, clip]{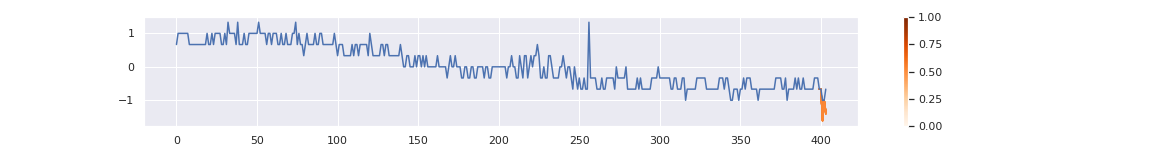}
    \includegraphics[width=1.3\textwidth, trim =4cm 0cm 7cm 0cm, clip]{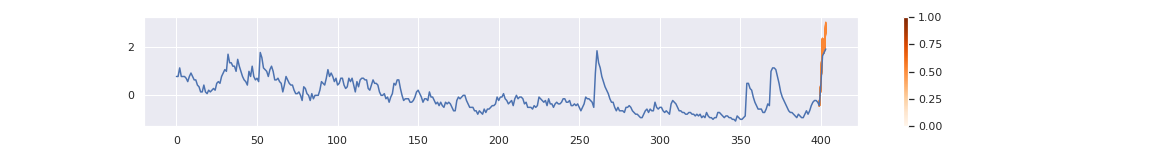}
    \includegraphics[width=1.3\textwidth, trim =4cm 0cm 7cm 0cm, clip]{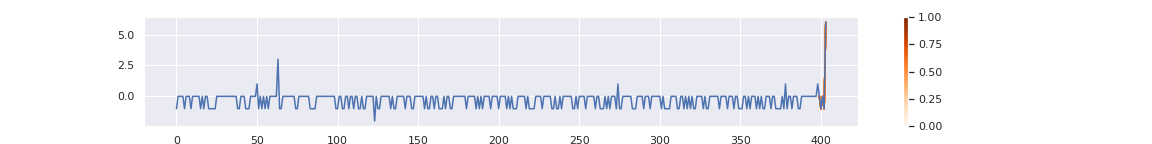}
    \includegraphics[width=1.3\textwidth, trim =4cm 0cm 7cm 0cm, clip]{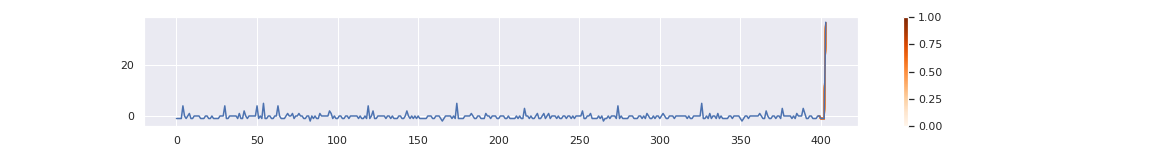}
    \subcaption{ICE}
    \label{fig:smd_icod}
\end{subfigure}
\hfill
\begin{subfigure}[b]{.45\textwidth}
     \centering
    \includegraphics[width=1.3\textwidth, trim =4cm 0cm 7cm 0cm, clip]{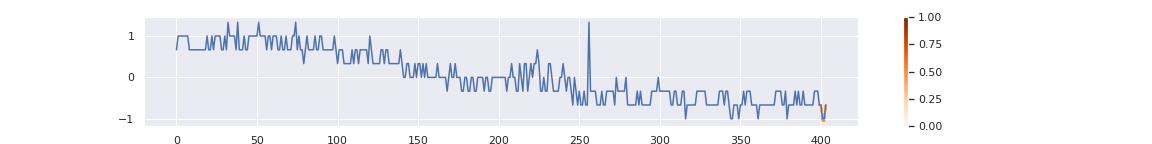}
    \includegraphics[width=1.3\textwidth, trim =4cm 0cm 7cm 0cm, clip]{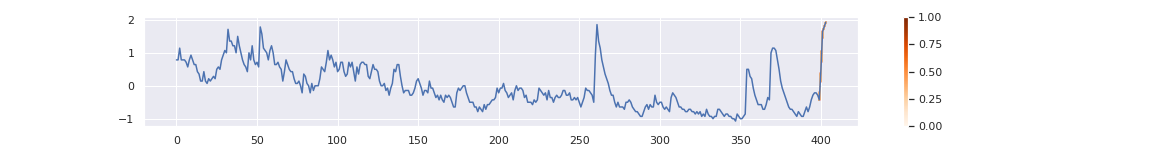}
    \includegraphics[width=1.3\textwidth, trim =4cm 0cm 7cm 0cm, clip]{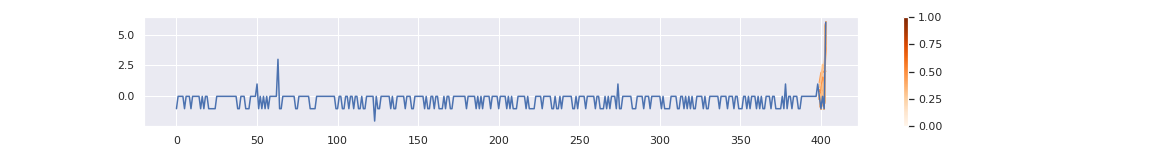}
    \includegraphics[width=1.3\textwidth, trim =4cm 0cm 7cm 0cm, clip]{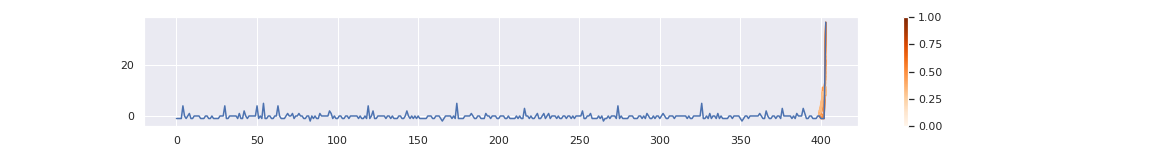}
    \subcaption{Sparse ICE}
    \label{fig:smd_icod_sparse}
\end{subfigure}
    \caption{Counterfactual explanation obtained with ICE (\ref{fig:smd_icod}) and the sparse variant (\ref{fig:smd_icod_sparse}). The different rows correspond respectively to the first, third, ninth and twelfth dimensions of a subsequence in the SMD dataset. Amongst them, only the fourth two (twelfth dimension) contains an anomalous observation in the last timestamp of the displayed window, detected by the NCAD model. While ICE (\ref{fig:smd_icod}) modifies all the plotted dimensions, Sparse ICE only perturbs the third and fourth (i.e., the ninth and twelfth dimension).} 
    \label{fig:comparison_icod_sparse}
\end{figure}

\begin{figure}
    \centering
    \begin{subfigure}[b]{.48\textwidth}
     \centering
    \includegraphics[width=\linewidth, trim={1cm 0cm 2cm 0cm},clip]{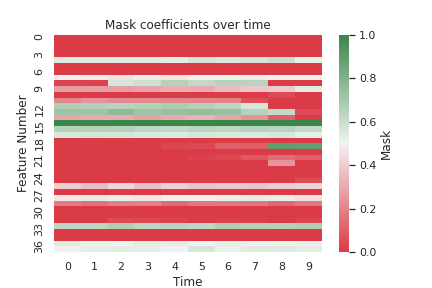}
    \subcaption{DPE}
    \label{fig:smd_mask}
    \end{subfigure}
    \hfill
    \begin{subfigure}[b]{.48\textwidth}
     \centering
     \includegraphics[width=\linewidth, trim={1cm 0cm 2cm 0cm},clip]{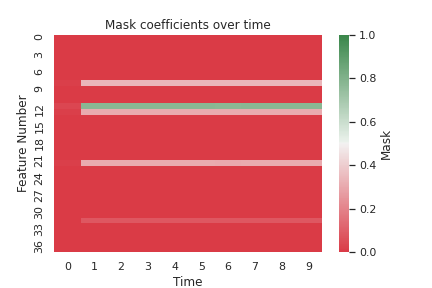}
    \subcaption{Sparse DPE}
    \label{fig:smd_mask_sparse}
    \end{subfigure}
    \caption{Perturbation maps of counterfactual examples in the explanations generated by DPE (left) and its sparse variant (right) on one anomaly in the SMD dataset detected by NCAD. We recall that the rows of each mask correspond to the different dimensions of the time series and the columns to the successive timestamps in the suspect window (see Section \ref{sec:methodology}). The color bars on the right sides of the maps indicate the values (between 0 and 1) of these maps along the time series features.}
    \label{fig:vis_masks}
\end{figure}

In this section, we report additional visualizations of our counterfactual explanations, as well as illustrations of the sparsity induced by the sparse variants of DPE and ICE. Figures \ref{fig:vis_kpi_ncad} and \ref{fig:vis_kpi_usad} are visualizations applied to the univariate datasets and respectively the NCAD and USAD. The advantage of Sparse ICE compared to the plain version ICE is shown in Figure \ref{fig:comparison_icod_sparse}, where only four channels of the multi-dimensional time series window are plotted. For this anomaly, only one of these dimensions contains an anomalous observation but the counterfactual explanation obtained with the plain ICE perturbs the four of them. In contrast, the Sparse ICE variant keeps two dimensions without anomalous features unchanged, leading to a more accurate and readable explanation on this particular anomaly. Similarly, Figure \ref{fig:vis_masks} shows two perturbation maps corresponding to examples generated by DPE and its sparse variant. While the plain DPE produces \emph{globally} sparse maps (i.e., in the temporal and dimensional features), Sparse DPE is sparse in dimensions, leading to perturbed examples with few modified channels.

\section{Illustration of the hyperparameters selection}\label{app:hps}

\begin{figure}
    \centering
    \includegraphics[width=0.49\linewidth]{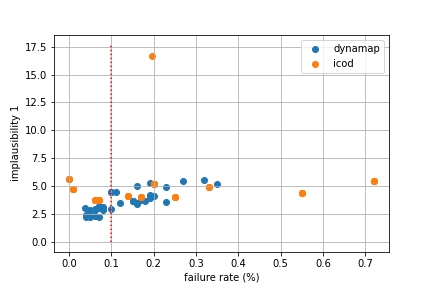}
    \includegraphics[width=0.49\linewidth]{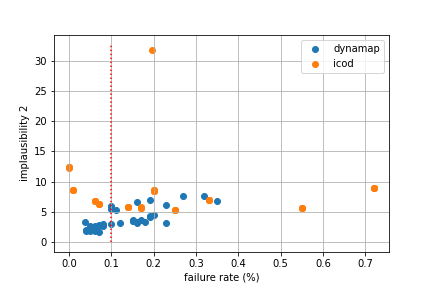}
    \includegraphics[width=0.49\linewidth]{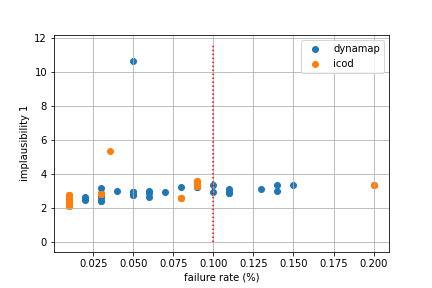}
    \includegraphics[width=0.49\linewidth]{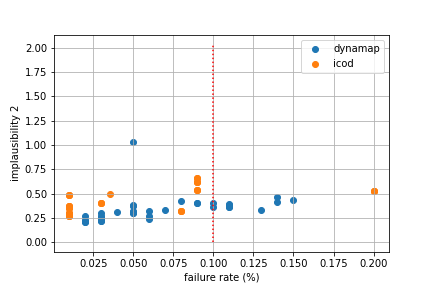}
    \caption{Implausibility measures 1 (left column) and 2 (right column) versus failures rates for different sets of hyperparameters of the ICE and DPE algorithms and their sparse variants applied to the NCAD (first row) and USAD (second row) models on a the KPI dataset. The metrics are computed over a validation set of 5 time series and the failure rate's threshold is 10\% (red dotted line).}
    \label{fig:hps_ncad_kpi}
\end{figure}

\begin{figure}
    \centering
    \includegraphics[width=0.49\linewidth]{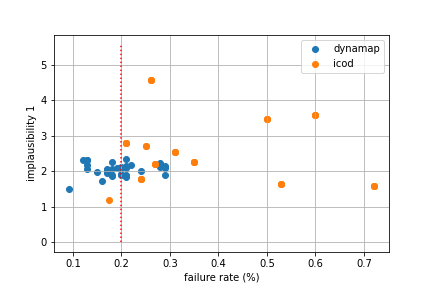}
    \includegraphics[width=0.49\linewidth]{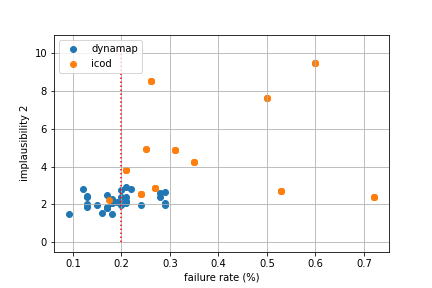}
    \includegraphics[width=0.49\linewidth]{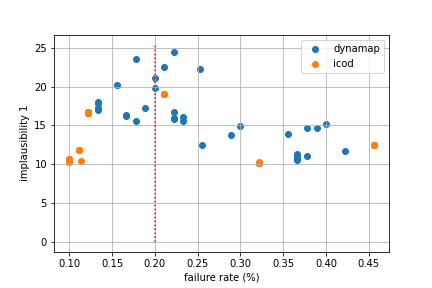}
    \includegraphics[width=0.49\linewidth]{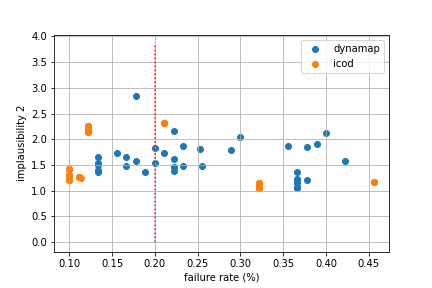}
    \caption{Implausibility measures 1 (left column) and 2 (right column) versus failures rates for different sets of hyperparameters of the ICE and DPE algorithms and their sparse variants applied to the NCAD (first row) and USAD (second row) models on a the Yahoo dataset. The metrics are computed over a validation set of 15 time series and the failure rate's threshold is 25\% (red dotted line).}
    \label{fig:hps_ncad_yahoo}
\end{figure}

\begin{figure}
    \centering
    \includegraphics[width=0.49\linewidth]{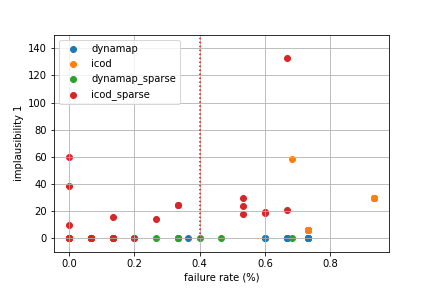}
    \includegraphics[width=0.49\linewidth]{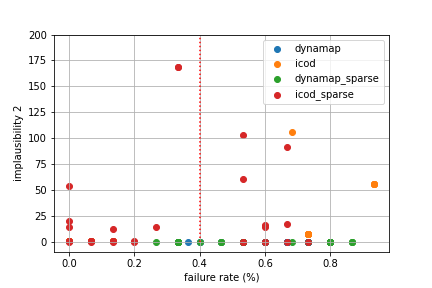}
    \includegraphics[width=0.49\linewidth]{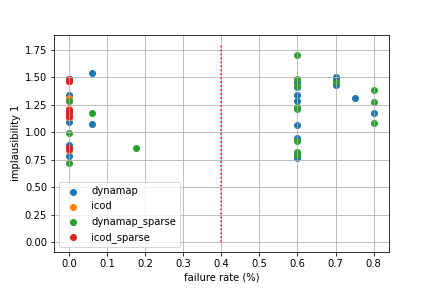}
    \includegraphics[width=0.49\linewidth]{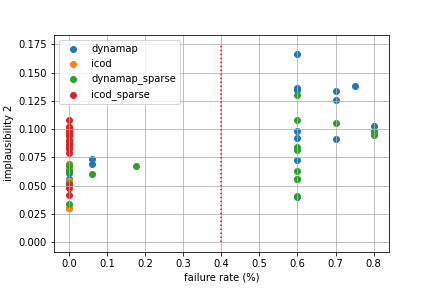}
    \caption{Implausibility measures 1 (left column) and 2 (right column) versus failures rates for different sets of hyperparameters of the ICE and DPE algorithms and their sparse variants applied to the NCAD (first row) and USAD (second row) models on a the SMAP dataset. The metrics are computed over a validation set of 40 time series and the failure rate's threshold is 25\% (red dotted line).}
    \label{fig:hps_ncad_smap}
\end{figure}

\begin{figure}
    \centering
    \includegraphics[width=0.49\linewidth]{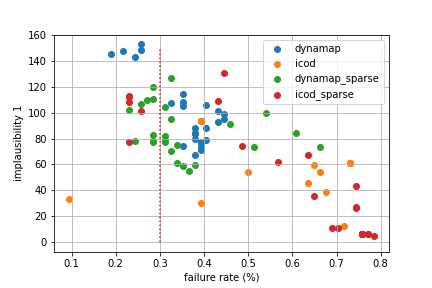}
    \includegraphics[width=0.49\linewidth]{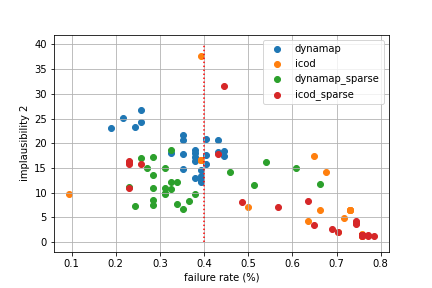}
    \includegraphics[width=0.49\linewidth]{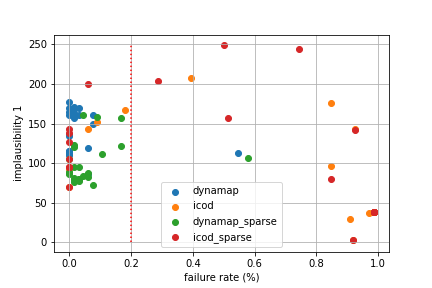}
    \includegraphics[width=0.49\linewidth]{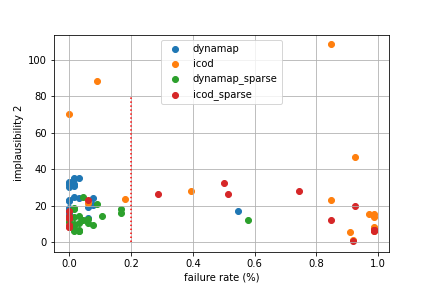}
    \caption{Implausibility measures 1 (left column) and 2 (right column) versus failures rates for different sets of hyperparameters of the ICE and DPE algorithms and their sparse variants applied to the NCAD (first row) and USAD (second row) models on a the SMD dataset. The metrics are computed over a validation set of 6 time series and the failure rate's threshold is 40\% for NCAD and 20\% for USAD (red dotted lines).}
    \label{fig:hps_ncad_smd}
\end{figure}

In this section, we illustrate the hyperparameters selection procedure for our gradient-based method. For each dataset and model, we run our algorithm with several configurations as described in Section \ref{sec:hp_selection} and select the final one using the failure rate and the Implausibility 1 metric. More precisely, we select a threshold of acceptable failure rate (e.g., 10\% or 20\%), then amongst the configurations achieving a lower value of the latter, we select the one with the lowest Implausibility 1 value. Figures \ref{fig:hps_ncad_kpi}, \ref{fig:hps_ncad_yahoo}, \ref{fig:hps_ncad_smd} and \ref{fig:hps_ncad_smap} show the values of these metrics for all explored configurations for each model and dataset. Lastly, in Tables \ref{tab:hp_dynamap}, \ref{tab:hp_icod}, \ref{tab:hp_dynamap_sparse} and \ref{tab:hp_icod_sparse}, we report the selected configurations for respectively DPE, ICE, Sparse DPE and Sparse ICE on the benchmark datasets. Besides, the hyperparameters of the gradient-free approach can be found in Table \ref{tab:hp_forecast}.

\begin{table}
    \centering
    \begin{tabularx}{\textwidth}{ 
   >{\centering\arraybackslash}X 
   >{\centering\arraybackslash}X 
   >{\centering\arraybackslash}X 
   >{\centering\arraybackslash}X 
   >{\centering\arraybackslash}X 
   >{\centering\arraybackslash}X 
   >{\centering\arraybackslash}X  } %
          \hline
          Dataset & Model type & Number of layers & Hidden size & training epochs & learning rate & prediction length \\
         \hline
KPI &  FFNN &    1   &   32 &    100 &     0.001 & 10\\
Yahoo &  FFNN &    1   &   32 &    100 &     0.001 & 10 \\
SMD &  DeepVAR & 4 & 40 & 150 & 0.001 & 10 \\
SWaT &  DeepVAR & 4 & 40 & 150 & 0.001 & 10 \\
         \hline
    \end{tabularx}
    \caption{Hyperparameters of the Probabilistic Forecasting models used in the gradient-free approach on the four benchmark datasets.}
    \label{tab:hp_forecast}
\end{table}

\begin{table}
    \centering
    \begin{tabularx}{\textwidth}{ 
   >{\centering\arraybackslash}X 
   >{\centering\arraybackslash}X 
   >{\centering\arraybackslash}X 
   >{\centering\arraybackslash}X 
   >{\centering\arraybackslash}X
   >{\centering\arraybackslash}X  } %
          \hline
          Dataset & Perturbation & $\sigma_{max}$ &  learning rate &  $\lambda_2$ & $\lambda_T$ \\
         \hline
NCAD-KPI &     Gaussian blur &        3.0 &          0.01 &                0.01 &              0.1 \\
NCAD-Yahoo &     Gaussian blur &        10.0 &         0.01 &                0.001 &              0.1  \\
NCAD-SMD &     Gaussian blur &       20.0 &         0.01 &                0.0 &                              1.0 \\
NCAD-SMAP   & Gaussian blur & 10.0 & 0.01 & 1.0 & 1.0 \\
USAD-KPI &     Gaussian blur &        3.0 &       0.01 &              0.001 &              1.0  \\
USAD-Yahoo &     Gaussian blur &        10.0 &       0.01 &              0.001 &              1.0   \\
USAD-SMD & Gaussian blur & 20.0 & 0.1 & 0.001 & 0.01 \\
USAD-SMAP & Gaussian Blur & 20.0 & 0.01 & 0.01 & 0.1 \\
         \hline
    \end{tabularx}
    \caption{Hyperparameters of the DPE algorithm on the four benchmark datasets. }
    \label{tab:hp_dynamap}
\end{table}

\begin{table}
    \centering
    \begin{tabularx}{\textwidth}{ 
   >{\centering\arraybackslash}X 
   >{\centering\arraybackslash}X 
   >{\centering\arraybackslash}X 
   >{\centering\arraybackslash}X 
   >{\centering\arraybackslash}X  } %
          \hline
          Dataset  & learning rate & $\lambda_1$ & $\lambda_2$ & $\lambda_T$  \\
         \hline
NCAD-KPI &           0.1 &           0.01 &               0.01 &                   1.0 \\
NCAD-Yahoo &         0.1 &                 0.01 &               0.01 &                   1.0 \\
NCAD-SMD &      0.1 &                 0.01 &                 0.01 &                   0.1 \\
NCAD-SMAP     & 0.1 & 0.1 & 0.1 & 1.0 \\
USAD-KPI &         0.1 &               0.001 &               0.001 &                   1.0 \\
USAD-Yahoo & 0.1  & 0.001 & 0.001 & 1.0 \\
USAD-SMD & 1000.0 & 0.01 & 0.01 & 1.0 \\
USAD-SMAP & 1.0 & 0.001 & 0.001 & 1.0 \\
         \hline
    \end{tabularx}
    \caption{Hyperparameters of the ICE algorithm on the four benchmark datasets. }
    \label{tab:hp_icod}
\end{table}

\begin{table}
    \centering
    \begin{tabularx}{\textwidth}{ 
   >{\centering\arraybackslash}X 
   >{\centering\arraybackslash}X 
   >{\centering\arraybackslash}X 
   >{\centering\arraybackslash}X 
   >{\centering\arraybackslash}X 
   >{\centering\arraybackslash}X 
   >{\centering\arraybackslash}X  } %
          \hline
          Dataset & Perturbation & $\sigma_{max}$  & learning rate  & $\lambda_1$ & $\lambda_2$ & $\lambda_T$  \\
         \hline
NCAD-SMD &     Gaussian blur &       20.0 &              0.1 &                      0.01 &                       0.01 & 0.1  \\
NCAD-SMAP   & Gaussian Blur & 10.0 & 0.01 & 0.1 & 0.1 & 0.1  \\
USAD-SMD & Gaussian Blur & 20.0 & 0.01 & 0.01 & 0.01 & 1.0 \\
USAD-SMAP & Gaussian Blur & 20.0 & 0.1 & 0.01 & 0.01 & 0.1 \\
         \hline
    \end{tabularx}
    \caption{Hyperparameters of the Sparse DPE algorithm on the two benchmark multivariate datasets. }
    \label{tab:hp_dynamap_sparse}
\end{table}

\begin{table}
    \centering
    \begin{tabularx}{\textwidth}{ 
   >{\centering\arraybackslash}X 
  >{\centering\arraybackslash}X 
   >{\centering\arraybackslash}X 
   >{\centering\arraybackslash}X 
   >{\centering\arraybackslash}X  } %
          \hline
          Dataset  & learning rate & $\lambda_1$ & $\lambda_2$ & $\lambda_T$   \\
         \hline
NCAD-SMD & 0.1 &                       0.01 &                        0.01 &                         0.1 \\
NCAD-SMAP   & 0.1 & 0.1 & 0.1 & 1.0  \\
USAD-SMD & 10000.0 & 0.01 & 0.01 & 0.1  \\
USAD-SMAP & 1.0 & 0.001 & 0.001 & 1.0  \\
         \hline
    \end{tabularx}
    \caption{Hyperparameters of the Sparse ICE algorithm on the two benchmark multivariate datasets. }
    \label{tab:hp_icod_sparse}
\end{table}

\begin{table}
    \centering
    \begin{tabularx}{\textwidth}{ 
   >{\centering\arraybackslash}X 
   >{\centering\arraybackslash}X 
   >{\centering\arraybackslash}X 
   >{\centering\arraybackslash}X 
   >{\centering\arraybackslash}X 
   >{\centering\arraybackslash}X 
   >{\centering\arraybackslash}X 
   >{\centering\arraybackslash}X  } %
          \hline
          Variant & Perturbation & $\sigma_{max}$  & learning rate  & $\lambda_1$ & $\lambda_2$ & $\lambda_T$ & $N$ \\
         \hline
ICE &     -  &      -  &              0.1 &                      0.01 &                       0.01 & 0.01 & 100 \\
DPE   & Gaussian Blur & 3.0 & 0.01 & - & 0.1 & 0.01 & 100  \\
         \hline
    \end{tabularx}
    \caption{Default set of hyperparameters for our gradient-based counterfactual ensemble method. }
    \label{tab:hp_default}
\end{table}

\section{Sensitivity of the Diversity criterion to the learning rate parameter}\label{app:sensitivity}

\begin{figure}
    \centering
    \includegraphics[width=0.45\linewidth]{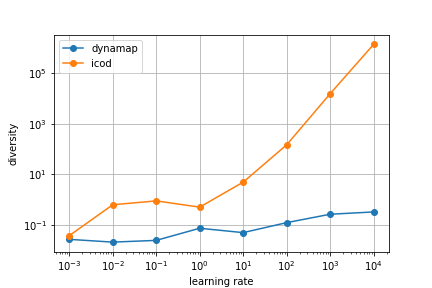}
    \includegraphics[width=0.45\linewidth]{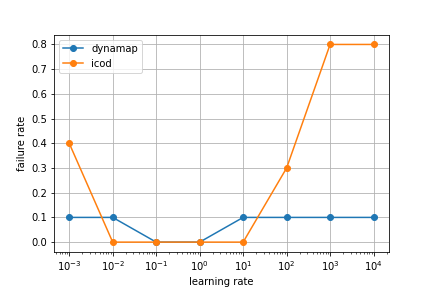}
    \caption{Diversity of the counterfactual ensemble (left) and failure rate of our counterfactual method (right) versus the learning rate of the SGD algorithm for the two variants of our method, ICE and DPE.}
    \label{fig:sensitivity_lr}
\end{figure}

In this section we report a small-scale study of the influence of the learning rate in the SGD algorithm on the Diversity metric, in our gradient-based approach. We evaluate the latter metric on 10 anomalies detected by the NCAD model in the KPI dataset, obtained with DPE and ICE with learning rates in the set $\{0.001, 0.01, 0.1, 1, 10, 100, 1000, 10000\}$. The other hyperparameters of our method are the same as in Section \ref{sec:exp_univariate}. Figure \ref{fig:sensitivity_lr} shows the evolution of the Diversity score (left panel) and failure rate (right panel) when the learning rate increases. We observe that the diversity is always higher for ICE than DPE, and dramatically increases when the learning rate is greater than 1 for the former. However, failure rate also skyrockets for high learning rates.

\end{document}